\documentclass[3p]{elsarticle}

\usepackage{amsmath,amssymb,upgreek}

\usepackage[utf8x]{inputenc}

\usepackage{nameref,hyperref}

\usepackage[]{lineno}

\usepackage[table]{xcolor}

\usepackage{array}

\usepackage{lastpage,fancyhdr,graphicx}

\usepackage{amsmath}
\usepackage{amsthm}
\usepackage{amsfonts}
\usepackage{amssymb}
\usepackage{graphicx}
\usepackage{mathalfa}
\usepackage{psfrag}
\usepackage[justification=centerfirst]{subfig}

\newcommand{\one}{($i$) }
\newcommand{\two}{($ii$) }

\begin{document}

\title{Random Feedback Alignment Algorithms to train Neural Networks: Why do they Align?} 
\author{Dominique Chu\corref{cor1}}
\cortext[cor1]{Corresponding author}
\ead{D.f.chu@kent.ac.uk}

 \author{Florian Bacho}
\ead{F.Bacho@kent.ac.uk}

\address{
School of Computing, CEMS, University of Kent, CT2 7NF, Canterbury, UK
}

\begin{abstract}
Feedback alignment algorithms are an alternative  to backpropagation to train neural networks, whereby  some of the partial derivatives that are required to compute the gradient are replaced by random terms. This essentially transforms the update rule into a random walk in weight space.  Surprisingly, learning still works with those algorithms, including training of deep neural networks.  This is generally attributed to  an alignment of the update of the random walker with  the true gradient --- the eponymous gradient alignment --- which drives an approximate gradient descend. The mechanism that leads to this alignment remains unclear, however.  In this paper, we use   mathematical reasoning and simulations to investigate gradient alignment.   We observe that the feedback alignment update rule has fixed points, which correspond to extrema of the loss function. We show that gradient alignment is a stability criterion for those fixed points.  It is only a necessary criterion for algorithm performance. Experimentally, we demonstrate that  high levels of gradient alignment can lead to poor algorithm performance and that the alignment is not always driving the gradient descend.
\end{abstract}


\maketitle

\section{Introduction}

The backpropagation algorithm (BP) \cite{error_backpropagation}  underpins a good part of modern neural network (NN) based AI. BP-based training algorithms continue to be the state of the art in many areas of machine learning ranging from benchmark problems such as the MNIST dataset\cite{mnist} to the most recent transformer-based architectures \cite{dfa_scales_to_modern_dl}.      While its success is undeniable, BP has some disadvantages. The main one is that BP is computationally expensive. It is so in  two ways. Firstly,  BP  has high requirements on  pure processing power. It also requires  sequential processing of layers during both the forward and backward pass, limiting its scope for parallelisation. This is sometimes called {\em backward locking} \cite{decoupled_parallel_bp}. 
\par
Secondly, BP is biologically implausible. One issue is that  for every neuronal feedforward connection, BP would require neurons to also have a symmetric feedback connection with the same weight. This has never been observed. From a purely machine learning point of view, the lack of biological plausibility may not be too concerning, since  the aim of applied AI is more often performance, rather than neuroscientific realism. However, there is a sense in which biological plausibility becomes a real concern after all: The update of any particular connection weight in a neural network requires global information about the entire network. This entails intense data processing needs \cite{famove}, which in turn leads to  high energy consumption \cite{sparse_dfa,dfa_cnn_online_learning_processor}. The electricity consumption required for  the training of large scale NNs is a barrier to adoption and environmentally unsustainable and is  widely recognised as problematic \cite{energyefficient1,energyefficient2}.  BP  is also not compatible with  neuromorphic  hardware platforms \cite{event_driven_random_bp_neuromorphic}, such as Loihi \cite{loihi} or SpiNNaker \cite{spinnaker}. 
\par
In the light of this, there has been some recent interest in alternatives to BP  that alleviate these issues \cite{forwardforward}. One particularly intriguing example are  {\em random feedback alignment} (FA) algorithms \cite{random_fa}. The basic FA algorithm  is just  BP with  the symmetric feedback weights replaced by a randomly chosen, but fixed, feedback matrix.  A variant of FA is {\em direct feedback alignment} (DFA) \cite{schluessel}, which   bypasses  backpropagation through layers and transmits the error directly to weights from the output layer via appropriately chosen feedback matrices. This enables  layer-wise parallel updates of NNs. Furthermore,  training no longer requires  global knowledge of the entire network, which makes it amenable to implementation on neuromorphic hardware.   FA and DFA  have been found to perform surprisingly well on a number of benchmark problems \cite{fabench,faspike}. Recently, it has been reported that they even work on  large-scale architectures such as transformers \cite{krzkrz}, often reaching performances that are comparable, albeit not exceeding, those of BP-based algorithms.  Both algorithms do not work well on  convolutional neural networks \cite{krzkrz}.
\par
FA algorithms replace  partial derivatives in the gradient computation  by random matrices. Mathematically, the  resulting update will no longer be a gradient of the loss function, but must be expected to be orthogonal to the gradient. It is therefore, at a first glance,  surprising that DFA and FA work at all.     A key insight into why they work was already given in  the original paper by Lillicrap \cite{random_fa} who showed that the update direction of FA is not orthogonal to the gradient  after all. They observed so-called {\em weight alignment}, whereby the weights of the network align with (i.e.~point into approximately the same direction as)  the feedback matrices and {\em gradient  alignment} where the updates of the FA algorithm align  with the gradient as computed by BP. They conjectured that this alignment drives  the approximate gradient descent of FA.
\par
    A mechanism that could lead to this alignment  was suggested by Refinetti  and co-workers \cite{refinetti}. They   modelled a linear  two-layer network using  a student-teacher setup  based on an  approach by Saad and Solla \cite{sollte}.  This  showed   that, at least in their setup,  when starting from  initially zero weights,  the weight update is  in the direction of the feedback matrix, leading to weight alignment and consequently gradient alignment. A corollary of their results  is  the prediction that  alignment is particularly  strong when the weights are initially vanishing. 
 Another important theoretical contribution is  by Nøkland \cite{schluessel} who formulated a stability criterion for DFA. 
\par
The above results were obtained using mathematically rigorous methods, but also rely on  restrictive simplifying assumptions (e.g.~linear networks in a student-teacher setup), which may or may not be relevant for realistic NNs. There is therefore a need to understand how FA operates in unrestricted NN, and whether the insights derived from simplified setups remain valid.  The aim of the present contribution is to shed more light onto why FA works. To do this, we will complement existing approaches and view FA as a random walk \cite{randomwalks}, or more specifically a spatially inhomogeneous random walk in   continuous weight space where   the distribution of jump lengths and directions varies according to  the position of the walker.  In the present case, the update is entirely determined by the FA update rule and the distribution of training examples. The latter acts as the source of randomness.  
\par
We will show below that across weight space there are particular points, that is specific choices of weights, where the jump length vanishes. In a slight abuse of notation, we will refer to those as  {\em fixed points} of the random walk. As will become clear, these  correspond to local extrema of the loss function, and as such correspond to  valid solutions of the BP algorithm. If the random walker landed exactly on one of those, then it would remain there. However, typically these fixed points are  not stable under the FA update rule, that is they are not  attractors of the random walker. In this case, a walker initialised in the neighbourhood of the fixed point would move away from the fixed point.  As one  of the main contributions of this paper we will show that gradient alignment is   the condition for fixed points to be stable.  This  stability criterion is  different from the one derived by  Nøkland \cite{schluessel} who showed that under certain conditions gradient alignment can lead to loss minimisation. We show here that feedback alignment {\bf is} the stability criterion to first order approximation and that it can be derived in a general way without any simplifying assumptions.
\par
 Furthermore, we will also  show that gradient alignment while necessary for FA to find good solution, is not a sufficient criterion. Based on simulation results, we will conjecture that alignment is not a driving the approximate gradient descent, but rather is a side-product of a random walk that is attracted by local extrema of the loss function.  Finally, based on extensive simulations, we will also propose a model of  how  NN learning under FA works.

\section{Results}

\subsection{Notation and basic setup}

We will start by introducing the notation and the basic setup on which the remainder of this paper is based.   Throughout, we will consider a feedforward neural network (multi-layer perceptron) parametrised by some weights $\mathbf w$. The network takes the   vectorised input $\mathbf x$ and returns the  output vector $\mathbf m(\mathbf x; \mathbf w)$. When the input is irrelevant, then we will use the shorthand notation $\mathbf m(\mathbf w)$ to describe the neural network. We consider a network of  $L$ layers, where each layer  $1\leq l \leq L$ comprises  $n_l$ artificial neurons, whose output is a scalar  non-linear functions $f_i^{(l)}(\cdot)$, where the index $1\leq i\leq n_l$ labels the neuron to which this output belongs. The argument to those activation function is the pre-activation function  
		\begin{displaymath}
h_j^{(l)} := \sum_{i=1}^{n_{l-1}} w_{ji}^{(l)}f_i^{(l-1)}
		\end{displaymath}
		with $w_{ji}^{(l)}\in \mathbb R$ denoting the parameters (or ``weights'') of $h_j^{(l)}$. For convenience, we will write $x_i^{(l)} := f_i^{(l)}$ and in particular  $f_i^{(0)} :=  x_i$ is the input to the network.  Throughout this manuscript, we denote the  loss function by  $\mathcal L(\mathbf m(\mathbf x))$, and assume that it is to be minimised via gradient  descend, although all our conclusions will remain valid for gradient ascend problems. 
\par
Finally, we define the {\em alignment measure} of two vectors or two matrices $\mathbf a$ and $\mathbf b$. In the case of two matrices, this is computed by flattening the matrices in some way, for example by stacking the columns to obtain $\mathbf a'$ and $\mathbf b'$. The alignment measure is then computed  as the inner product of the vectors divided by their norms,
\begin{equation}
\label{alignemeneq}
{\mathbf a'\cdot\mathbf b'\over \lVert\mathbf a'\rVert \lVert\mathbf b'\rVert}.\tag{alignement measure}
\end{equation}
 The maximal value of the alignment is 1. This can be interpreted as $\mathbf a'$ and $\mathbf b'$ being  completely parallel. The minimal value is $-1$, which indicates that they are anti-parallel. In high dimensional spaces, two  randomly chosen matrices/vectors will typically have an alignment of 0, indicating that they are orthogonal to one another. 

\subsection{BP and FA algorithms}

 Using this notation, we can formulate  the BP update rule for layer $l$  of a feedforward multi-layer perceptron as 
\begin{equation}
\label{basicequation}
\Delta^\textrm{BP} w_{pq}^{(l)} :=  {\partial\mathcal L\over\partial f_i^{(L)}} {\partial f_i^{(L)} \over  \partial h_j^{(L)}  }  {\partial h_j^{(L)} \over  \partial f_l^{(L-1)}  } {\partial f_l^{(L-1)} \over \cdots}
\cdots   
{\cdots\over \partial f_k^{(l)}}
{\partial f_k^{(l)} \over  \partial h_s^{(l)}  }  {\partial h_s^{(l)} \over  \partial  w_{pq}^{(l)}}.
\end{equation}
Here (and in the following), we use the convention that repeated indices are summed over, that is  $a_ib_i := \sum_i a_ib_i$; note that this convention does not apply to the superscripts in parenthesis that indicate the layer.   
\par
Equation \ref{basicequation} can be evaluated, by noting that the function $f_i^{(l)}$ only depends on $h_i^{(l)}$, and furthermore for the common type of neural network, ${\partial h_i^{(l)} \over  \partial  w_{jk}^{(l)}} = \delta_{ij} f_k^{(l-1)}$, where $\delta_{ij}$ is $1$ if $i =j$ and $0$ otherwise. Thus, eq.~\ref{basicequation} reduces to
\begin{align}
\label{nextequation}
\Delta w_{pq}^{(l)}&= 
{\partial\mathcal L_i} \widetilde{B}_{ik}^{(l)} {\partial f_k^{(l)} \over  \partial h_s^{(l)}  }  {\partial h_s^{(l)} \over  \partial  w_{pq}^{(l)}} \nonumber\\
&= 
{\partial\mathcal L_i} \widetilde B_{ik}^{(l)} {\partial f_{ks}^{(l)}   }  {\delta_{ps}}f_q^{(l-1)} \nonumber\\
&={\partial\mathcal L_i} \widetilde   B_{ik}^{(l)} {\partial f_{kp}^{(l)}   }  f_q^{(l-1)}.
\end{align}
Here we abbreviated the partial derivative of the loss  by ${\partial\mathcal L_i}$, $\partial f_{ks}^{(l)} := \partial f_{k}^{(l)}/ \partial h_{s}^{(l)}$, and $\widetilde B_{ik}^{(l)}$ is a shorthand for the middle terms  of  the chain rule of eq.~\ref{basicequation}. Note that, $\partial f_{ks}^{(l)}$ is zero for $k\neq s$.  
%
%
%
%

FA is the same as BP, except that the terms ${\partial h_i^{(\cdot)} /  \partial f_j^{(\cdot)}}$ appearing  in  $\widetilde B_{ik}^{(l)}$  are replaced by  randomly chosen (but fixed) numbers $R_{ij}$, drawn from some user-determined distribution. This  leads to the partially random feedback matrices $\mathbf B$ with elements  $B_{ik}^{(l)}$. The  {\em FA pseudo-gradient} in layer $l$  is defined by 
 
\begin{equation}
\label{pseudogradient}
\Delta^{\textrm{FA}} w_{pq}^{(l)} 
:={\partial\mathcal L_i} B_{ik}^{(l)} {\partial f_{kp}^{(l)}   }  x_q^{(l)}.
\end{equation}
Note, that the rhs of the equation is not a  gradient of a a particular function. 
\par
 DFA is the same as FA, except that   the elements of the  matrix    $B_{ik}^{(l)}$  are  fixed but   chosen randomly.

\subsection{Deriving the stability criterion for FA}

The dynamics  induced by the FA-pseudo-gradient (eq.~\ref{pseudogradient}) constitutes a random walk in weight space. The randomness is introduced via the choice of the particular training example for the next weight update. Therefore,  FA is not a gradient descent/ascent algorithm, except of course in the  output layer which  follows a gradient to a local extremum of the loss function (in exactly the same way as BP).     
\par
We will now show that as a consequence of this, the FA pseudo-gradient update shares  with BP  a number of fixed points in weight space. These  correspond to local extrema of the loss function. Under certain conditions, FA will converge to those.  In order to understand the difference between the FA and BP it is instructive to consider the update in the penultimate layer of the network (which has the simplest form). In the case of BP, this is:
\begin{align}
\Delta^\textrm{BP} w_{pq}^{(l)} &=  
{\partial\mathcal L\over\partial f_i^{(L)}} {\partial f_i^{(L)} \over  \partial h_j^{(L)}  }  {\partial h_j^{(L)} \over  \partial f_l^{(L-1)}  } {\partial f_l^{(L-1)} \over \partial h_k^{(L-1)}}  {\partial h_k^{(L-1)} \over  \partial  w_{pq}^{(L-1)}} 
\nonumber \\ &=
{\partial\mathcal L\over\partial f_i^{(L)}} {\partial f_i^{(L)} \over  \partial h_j^{(L)}  }  { w_{jl}^{(L)}} {\partial f_l^{(L-1)} \over \partial h_k^{(L-1)}}  {\partial h_k^{(L-1)} \over  \partial  w_{pq}^{(L-1)}} \tag{gradient}
\end{align}
The corresponding expression in the case of FA is then:
\begin{equation}
\label{faupdate}
\Delta^\textrm{FA} w_{pq}^{(l)} =  
{\partial\mathcal L\over\partial f_i^{(L)}} {\partial f_i^{(L)} \over  \partial h_j^{(L)}  }  {  R_{jl}^{(L)}} {\partial f_l^{(L-1)} \over \partial h_k^{(L-1)}}  {\partial h_k^{(L-1)} \over  \partial  w_{pq}^{(L-1)}} \tag{FA pseudo-gradient}
\end{equation}
where $\mathbf R$ is a randomly chosen, but fixed matrix. 
\par
\begin{figure}
\centering
\subfloat[][]{\includegraphics[width=0.45\textwidth]{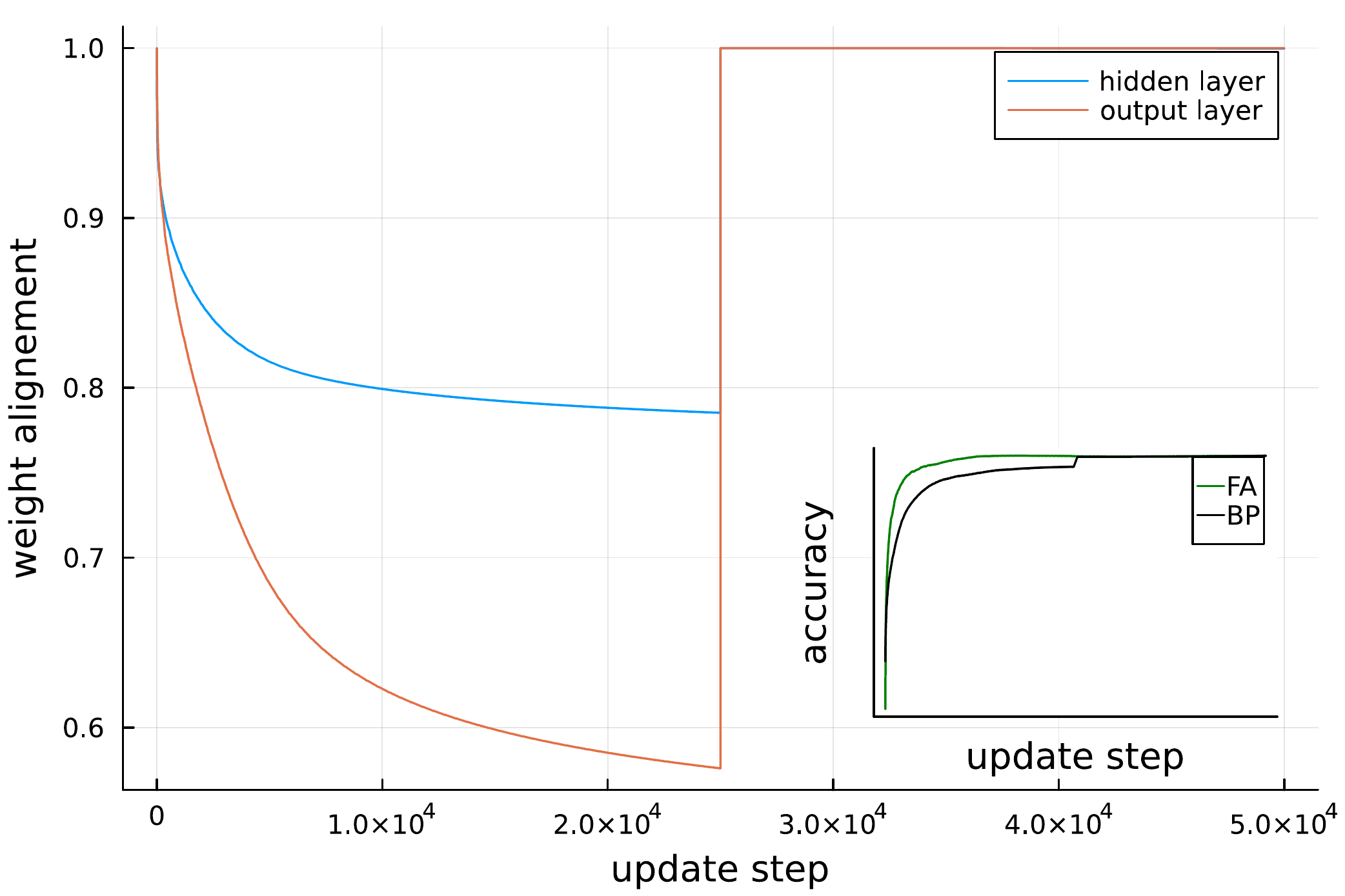}\label{swap}}
\subfloat[][]{\includegraphics[width=0.45\textwidth]{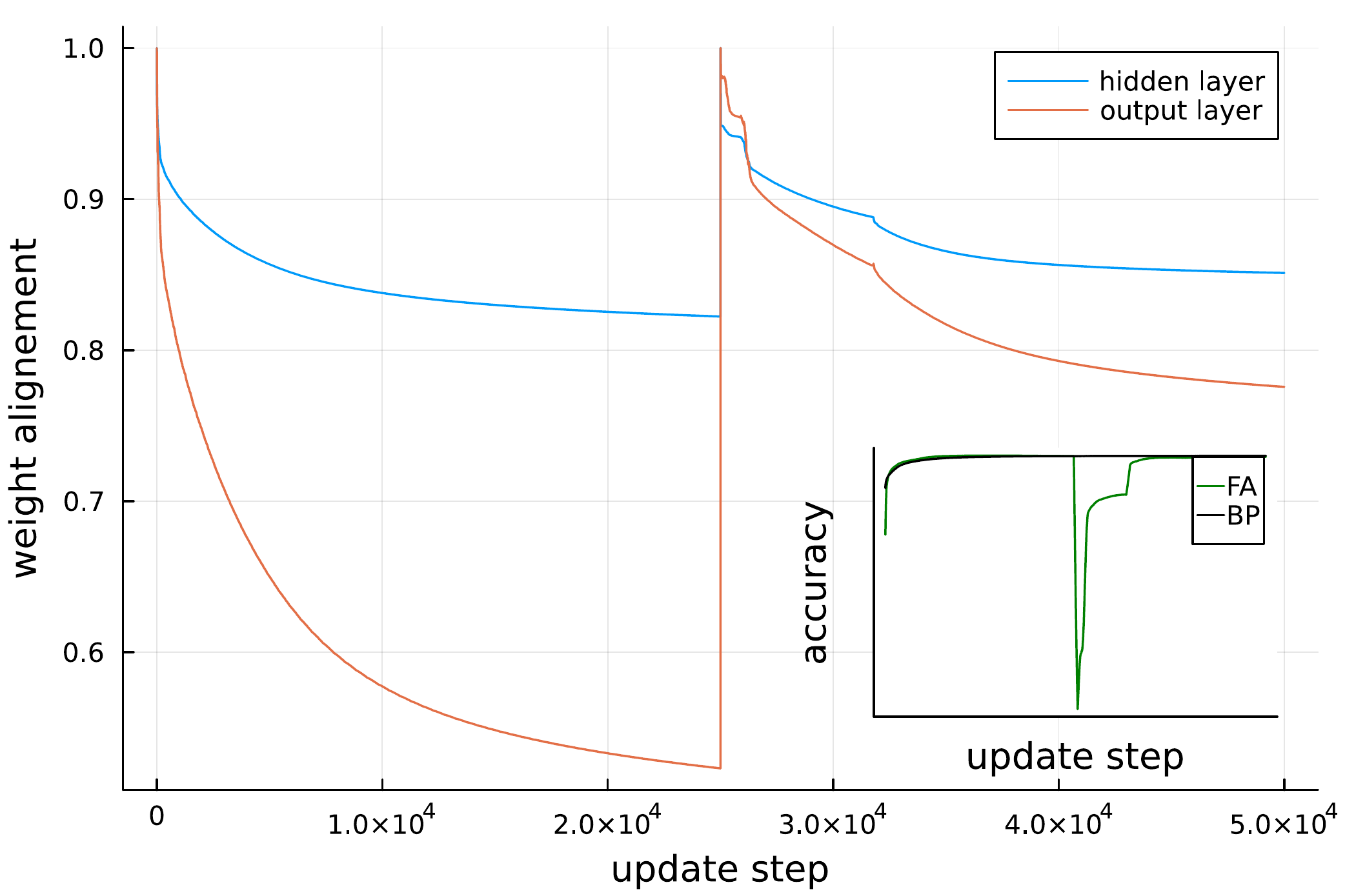}\label{swap2}}
\caption{ FA and BP were trained on the same sequence of examples from  MNIST starting from the same initial conditions. The graphs show the weight alignment over time between FA and BP for the hidden and output layer. A value of 1 means that FA and BP have the same weights up to a constant factor. \protect\subref{swap}  Initially, the two algorithms diverge. After 25000 update steps the weights of the FA network were transferred to the network trained using BP. Following this,  they remain aligned, demonstrating that the solution found by FA remains stable under BP. The inset shows the accuracy for both for reference.\protect\subref{swap2} Same, but at 25000 the FA takes the weight found by BP. This is not stable and the two solutions diverge quickly following the swap. Note how the accuracy of the FA drops immediately following the weight swap, indicating that it is repelled from the loss extremum.}
\label{swaps}
\end{figure} 
\par
We see from  the above equations  that for BP the gradient vanishes at each layer when the the derivative of the loss $\left(\partial\mathcal L/ \partial f_i^{(L)}\right) \left(\partial f_i^{(L)} /  \partial h_j^{(L)}\right) $ vanishes for all indices $j$.  If the weight matrices are full rank, then this is indeed the only way for the gradient to vanish. We observe that  random matrix $\mathbf R$ will be maximal rank  for as long as elements are chosen {\em iid}.  Its nullspace vanishes and local extrema of the loss function are therefore also points where the update of the FA pseudo-gradient vanishes. For as long as $\mathbf R$ is full rank, then all fixed points of the FA pseudo-gradient will be local extrema of the loss function.  As a consequence,   the fixed points of  FA will be local extrema of the loss function, and local attractors under the  BP.  
This means, that the optimiser in the output layer pushes the entire network, to fixed points of the FA pseudo-gradient.  Note that these fixed points need not be attractors of the  FA pseudo-gradient. When initialised close to such a fixed point, it is conceivable that FA moves away from the neighbourhood of the fixed point. Indeed, it can be easily seen that this happens (see fig.~\ref{swaps} for an example).
\par
The question is now whether or not there are fixed points that are attractive under the FA pseudo-gradient.  The condition for stability is that under the  FA update $\Delta^\textrm{FA}\mathbf w$ the loss does not increase, such that 
\begin{equation}
\mathcal L\left(\mathbf m\left(\mathbf w\right)\right) - \mathcal L\left(\mathbf m\left(\mathbf w  - \Delta^\textrm{FA}\mathbf w^{(l)} \right)\right) \geq 0.
\end{equation}
Here, we suppressed the label superscripts for clarity and wrote $\mathbf w$ instead of $\mathbf w^{(l)}$.  Assuming the weight update is a small one, we can now  expand to first order  and obtain
\begin{equation}
\mathbf m\left(\mathbf w - \Delta^\textrm{FA}\mathbf w \right) 
\approx 
\mathbf m\left(\mathbf w\right) - \mathbf m'\left(\mathbf w\right)\Delta^\textrm{FA}\mathbf w,
\end{equation}
where    $\mathbf m'$ is a   three-dimensional matrix with elements $m'_{ijk} := \partial m_i/\partial w_{jk}$.  We can now further expand the  loss function to first order to obtain
\begin{equation}
\mathcal L\left(m\left(\mathbf w  \right)\right)  - \mathcal L\left(\mathbf m\left(\mathbf w - \Delta^\textrm{FA}\mathbf w \right)\right)
\approx 
{\partial \mathcal L\over \partial \mathbf m} \mathbf m'(\mathbf w  ) \Delta^\textrm{FA} \mathbf w.
\end{equation}
Thus the stability criterion becomes, in  first order approximation
\begin{equation}
\label{alignder}
{\partial \mathcal L\over \partial \mathbf m} \mathbf m'(\mathbf w  ) \Delta^\textrm{FA} \mathbf w\geq 0
\end{equation}
We observe that ${\partial \mathcal L\over \partial \mathbf m} \mathbf m'(\mathbf w  )$ is just $\Delta^\textrm{BP} \mathbf w$, and $\Delta^\textrm{FA} \mathbf w$ is  the FA  pseudo-gradient. Furthermore, we see that the rhs of eq.~\ref{alignder} is  up to normalisation the alignment measure of the FA update with the BP gradient. Thus, eq.~\ref{alignder} formulates a necessary  condition for the stability of a local extremum with respect to the update by the FA pseudo-gradient. This stability criterion states that   the alignment measure needs to be positive.
\begin{equation}
{
\left(\Delta \mathbf w\right)^\textrm{BP} \cdot \left(\Delta \mathbf w\right)^\textrm{FA}
\over
\lVert\Delta \mathbf w\rVert^\textrm{BP} \cdot \lVert\Delta \mathbf w\rVert^\textrm{FA}
}
\geq 0
\label{alignemnetcriterion}
\end{equation}
Put differently, we can now say that gradient alignment is a stability criterion for FA.

\subsection{A conjectured mechanism}

The FA pseudo-random update rule is a spatially inhomogeneous random walk, whose update steps depend both on the input $\mathbf x$  to the network and the current weights $\mathbf w$. A priori, there cannot be an expectation that the walker moves along a trajectory that reduces the loss because the FA pseudo-gradient is not a gradient, except in the output layer. 
\par
Based on the form of the FA pseudo-gradient, we can still  conjecture a mechanism for the walker to find a local extremum of the loss function. 
\begin{itemize}
\item
Initially, the walker moves randomly through parameter space in the input and hidden layers. 
\item
In the output layer,  the error  is minimised, driving the derivative of the loss function  $\partial\mathcal L_i$ to zero, and hence (as discussed above) towards a fixed point of the FA pseudo-gradient. 
\item
The relevant fixed point may be unstable under the FA pseudo-gradient update rule. In this case,  the  FA random walker in the input and hidden layers will have a systematic drift,  making it impossible for the gradient descent process in the output layer to settle on its local extremum. 
\item
Convergence to a fixed point will only happen when the trajectory towards this fixed point is compatible with the  stability criterion  for all layers. 
\end{itemize} 
  There is no guarantee  that such a compatible extremum is found. For example, the relevant loss extrema may not be accessible from the initial state, or there may be no extrema for which the stability criterion holds.   It could also be that the random walk gets stuck in a region with near-zero update step sizes.  For example,  the  local derivative ${\partial f_{ij}^{(l)}}$  vanishes  as weights go to infinity (see \ref{basicproperties} for a discussion of this). This prevents the random walker from exploring  the weight space.  

\subsection{Experiments}

In the following, we will try to understand the behaviour of the random walker experimentally by focussing on the particular example of a 3 layer feed-forward neural network trained on the MNIST problem. As we will see below, for this problem and our parameter settings, FA works reasonably well, which makes this a suitable example to gain some insights into how FA works. Note that  in what follows  the focus is on understanding FA. The NN used and the MNIST benchmark are chosen as illustrative  examples.  We  are not trying to optimise the performance of FA on this task, nor are we attempting to give novel approaches to solve MNIST. Instead, the sole aim of the   experiments below is to gain some new insights about FA by observing how it works. The example itself (MNIST) is convenient, because it is easy to solve, but the precise choice of example is to some  extent irrelevant, and others could have been used to obtain the same insights. 

\subsubsection{Alignment}

There are two meanings to alignment in FA.  \one Weight alignment and \two gradient alignment. The consensus view in the  literature on FA  is  that  FA (somehow) brings about weight alignment, which implies  gradient alignment.    The latter then  drives the FA towards extrema of the loss function. In that sense,  FA approximates BP. In this section, we will show some examples where the approach to the local extremum is not driven by weight alignment, at least not during some stages of the simulations. 
 
 \par
%
\begin{figure}
\centering
\subfloat[][]{\includegraphics[width=0.45\textwidth]{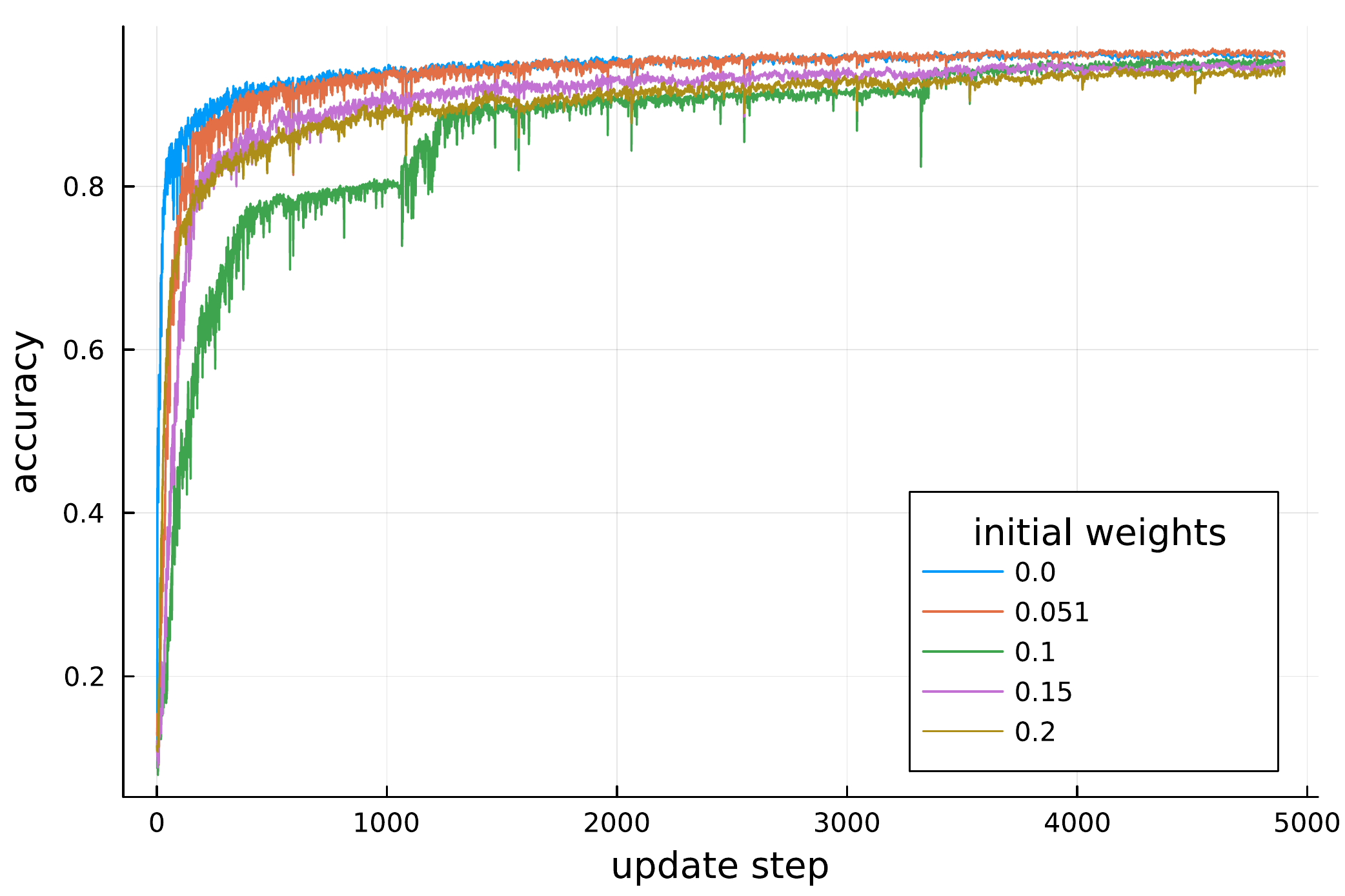}\label{alignementd}}
\subfloat[][]{\includegraphics[width=0.45\textwidth]{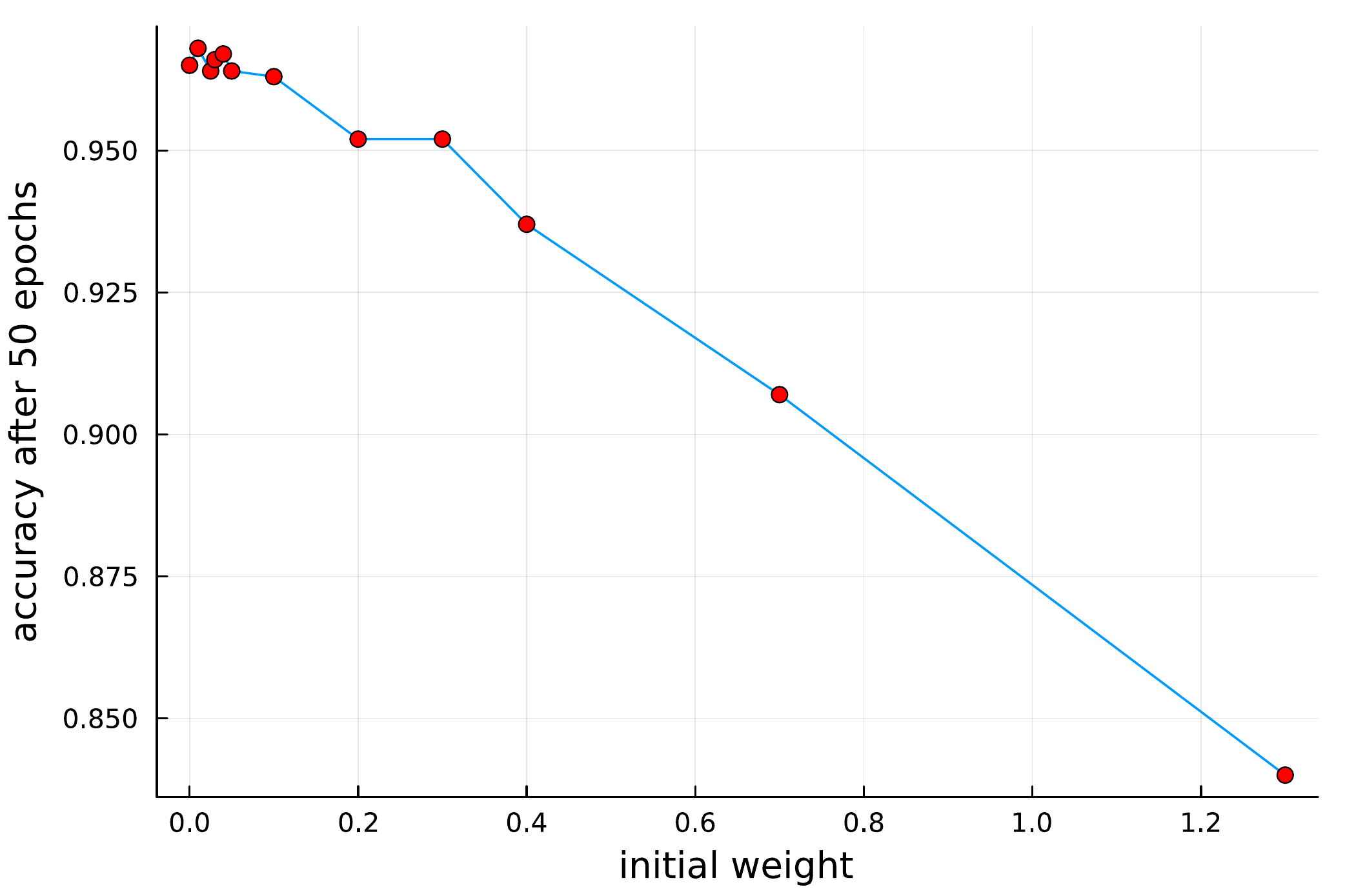}\label{alignementc}}\\
\subfloat[][]{\includegraphics[width=0.45\textwidth]{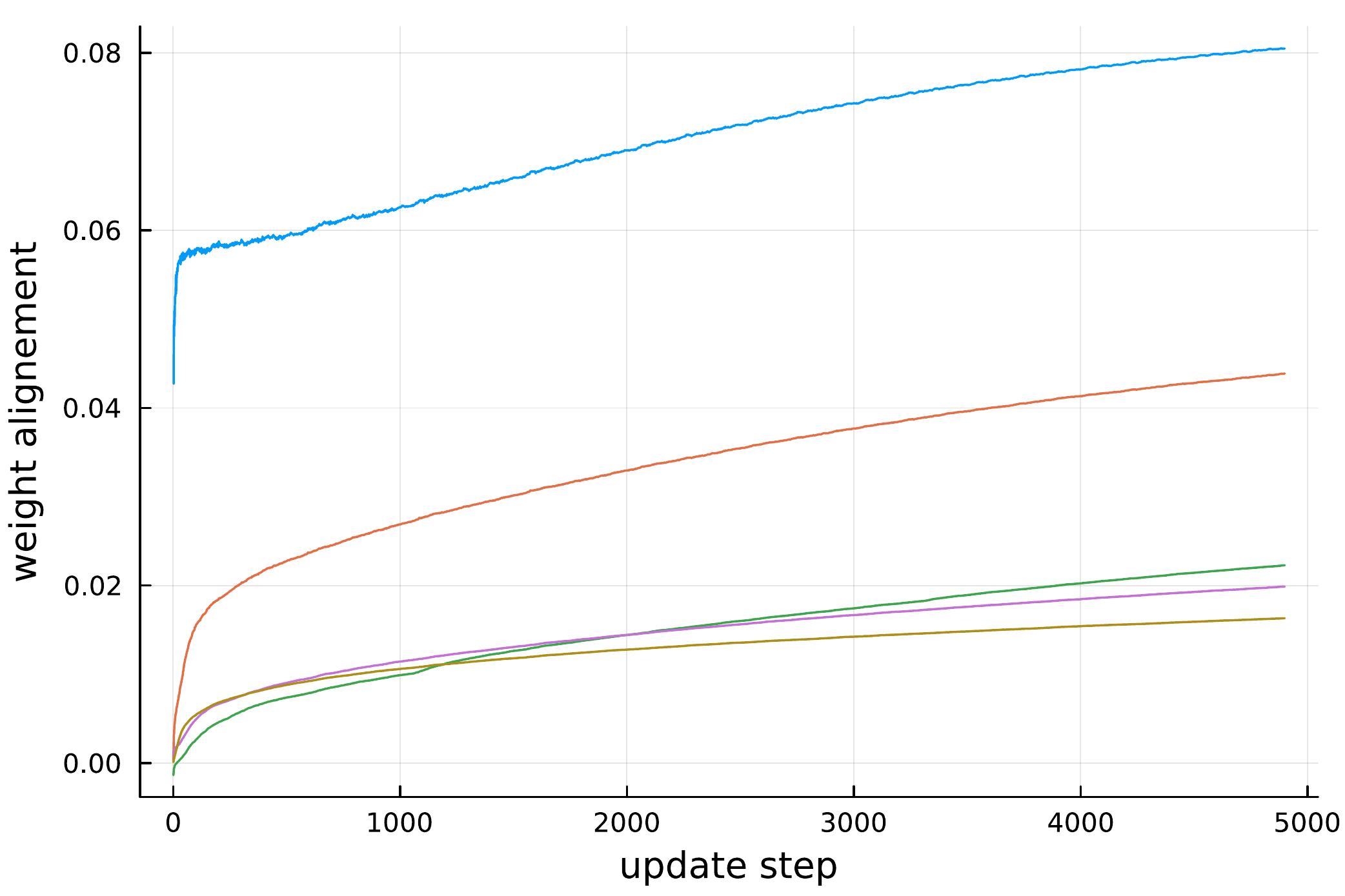}\label{alignementa}}
\subfloat[][]{\includegraphics[width=0.45\textwidth]{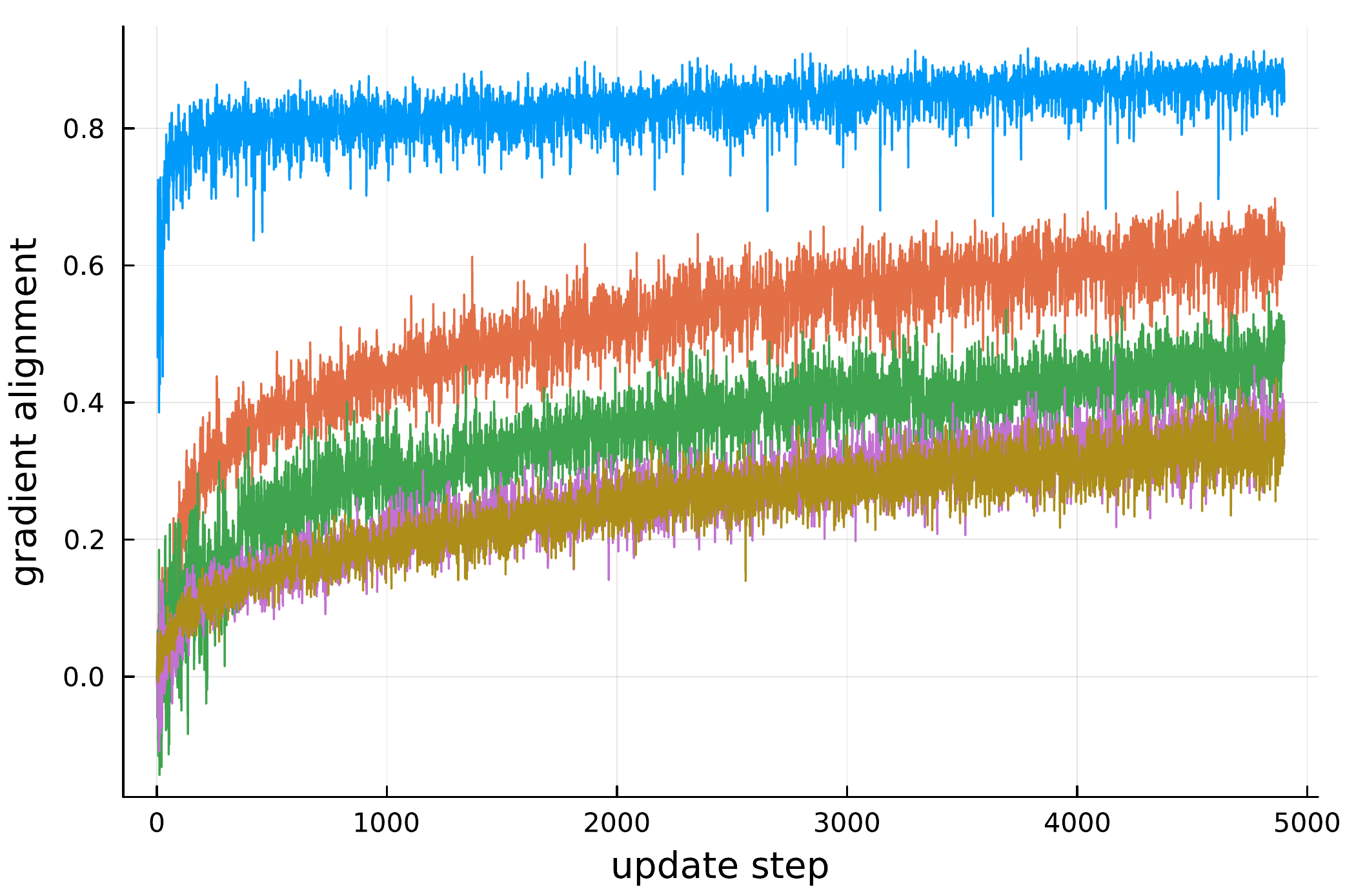}\label{alignementb}}
\caption{ The upper panel shows    \protect\subref{alignementd}  the accuracy as a function of the update step for altogether 10 epochs.   \protect\subref{alignementc} The accuracy after 50 epochs  as a function of the initial weights (see Methods for an explanation). 
 An approximate linear dependence is discernible. The lower panel shows the \protect\subref{alignementa} weight alignment and  \protect\subref{alignementb} gradient alignment for various initial weights as a function of updates. All results are averaged over 3 independent repetitions and used standard parameters (see Methods).  } 
\label{alignement}
\end{figure}
One of the key results by  Refinetti {\em et al} is that when starting from initially vanishing weights, the update of weights is in the direction of the feedback matrices. If this is true and gradient alignment drives FA learning, then one would expect     that small initial weights  lead to initially larger alignment than non-vanishing initial weights and   that  low initial weights  lead to faster convergence of FA to a good solution.     Our simulations are consistent with this.  Fig.~\ref{alignement} shows  that  there is higher weight alignment   (fig.~\ref{alignementa}) and gradient alignment (fig.~\ref{alignementb}) when   initial weights are small. Interestingly,   weight alignment remains  rather modest in comparison to the gradient alignment. Within 10 epochs, it  does not even reach a value of $0.1$.  Still, as expected, FA finds good solutions faster when starting with lower weights (fig.~\ref{alignementd}). This conclusion also holds in the long run.  Even after 50 epochs, the initial conditions matter for the achieved accuracy.  The higher the initial weights, the lower the accuracy  (see fig.~\ref{alignementc}).  
%
%
%
\begin{figure}
\centering
\subfloat[][]{\includegraphics[width=0.30\textwidth]{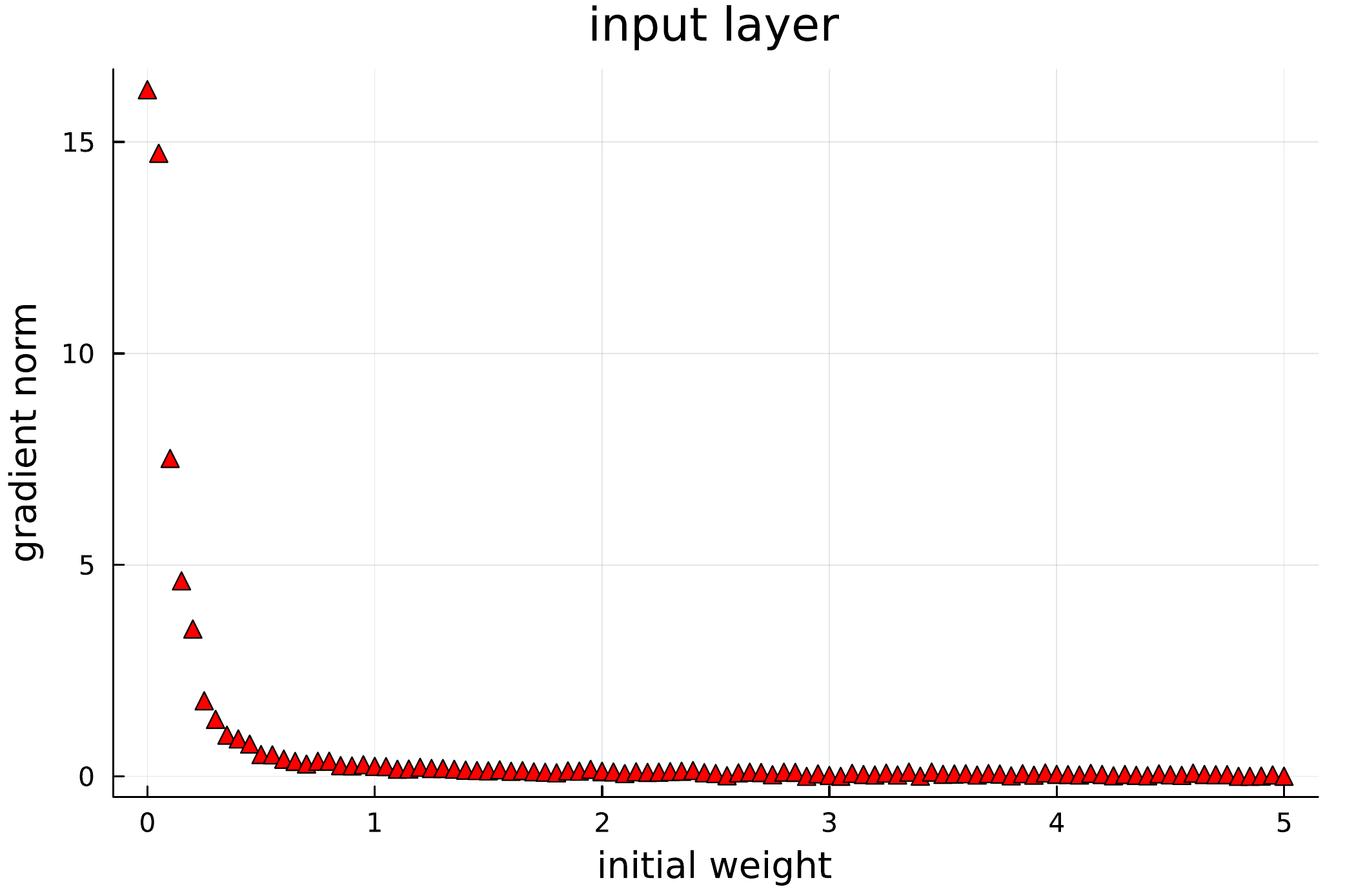}\label{initialgrada}}
\subfloat[][]{\includegraphics[width=0.30\textwidth]{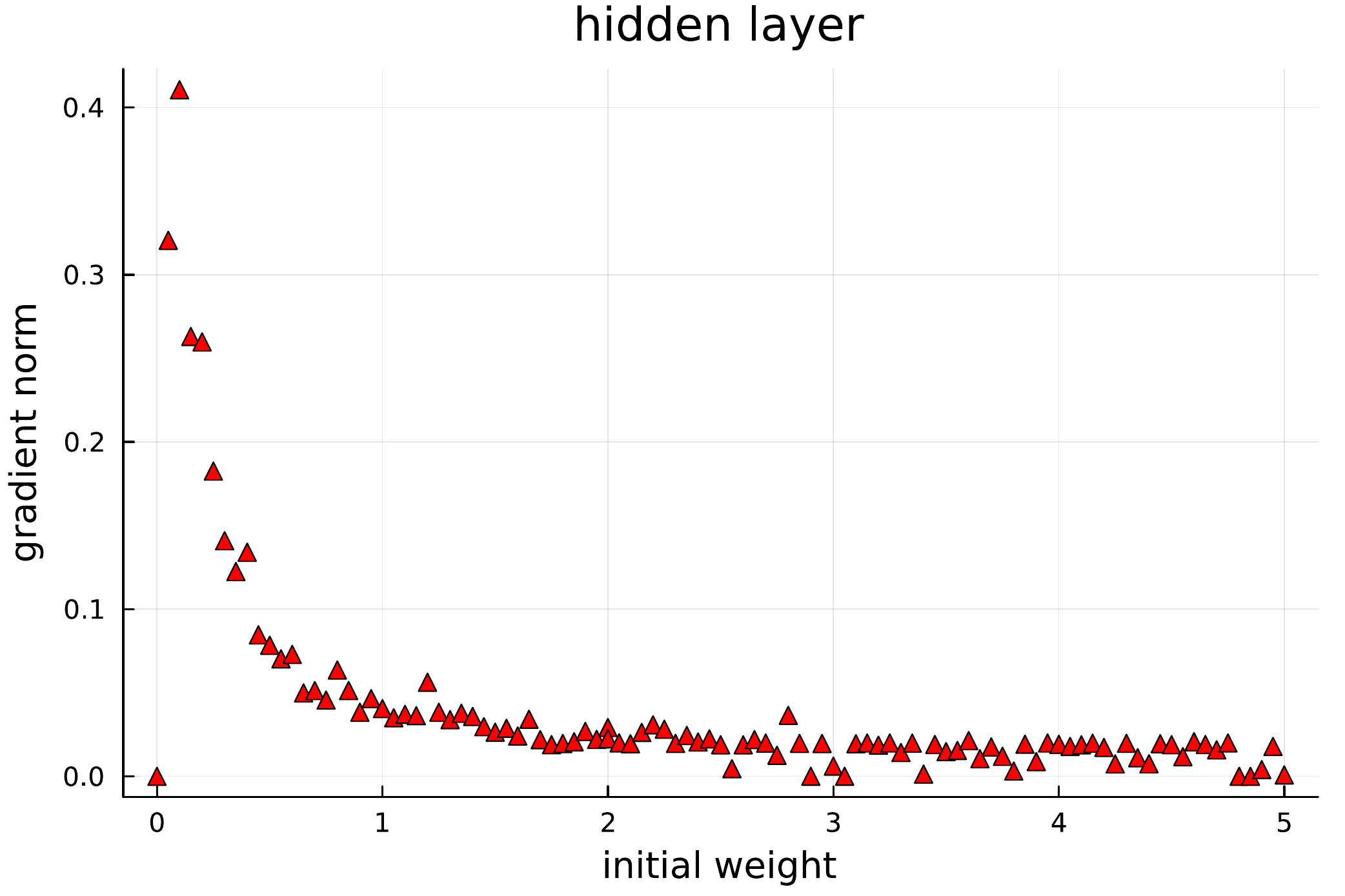}\label{initialgradb}}
\subfloat[][]{\includegraphics[width=0.30\textwidth]{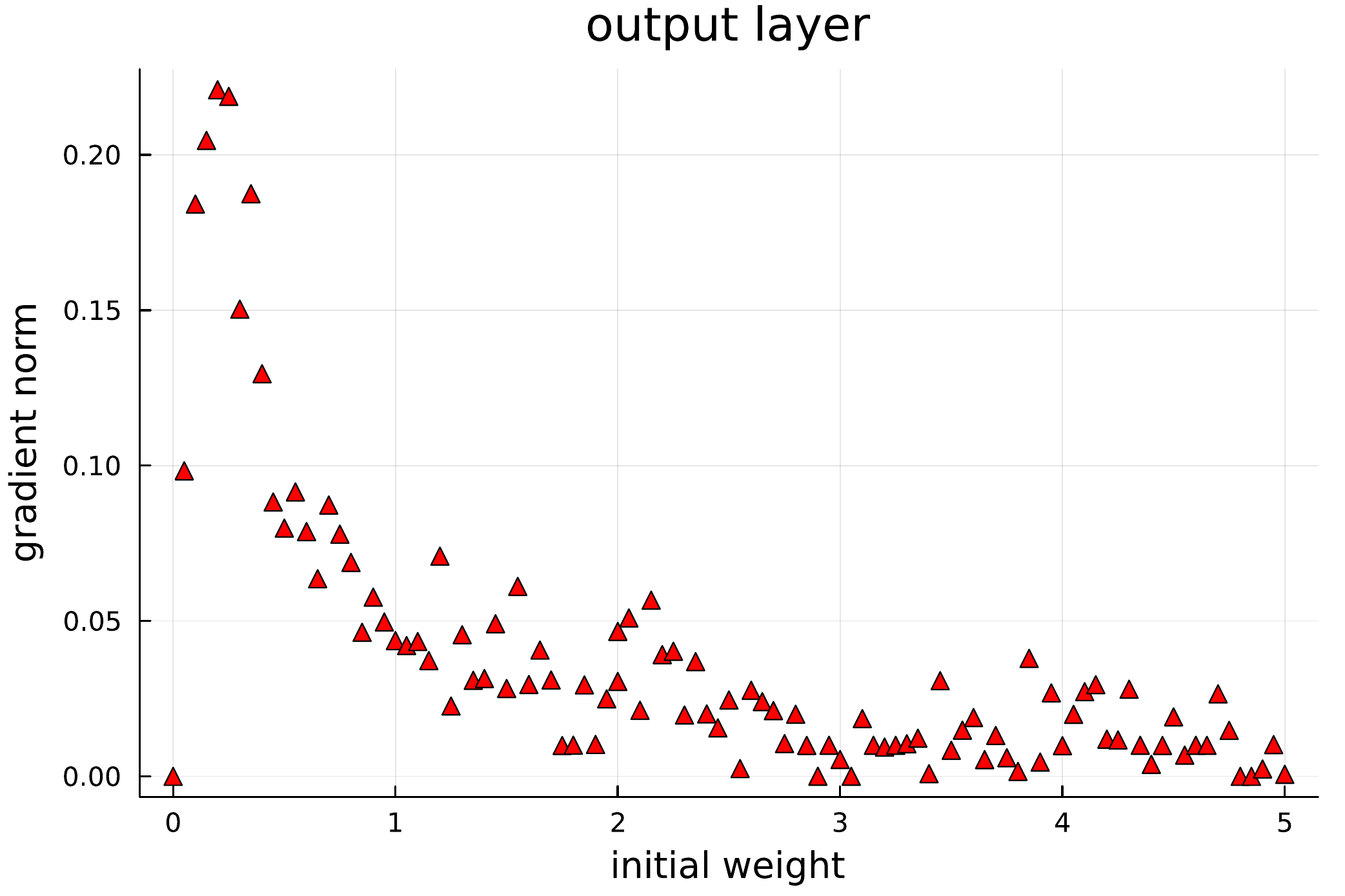}\label{initialgradc}}\\
\subfloat[][]{\includegraphics[width=0.30\textwidth]{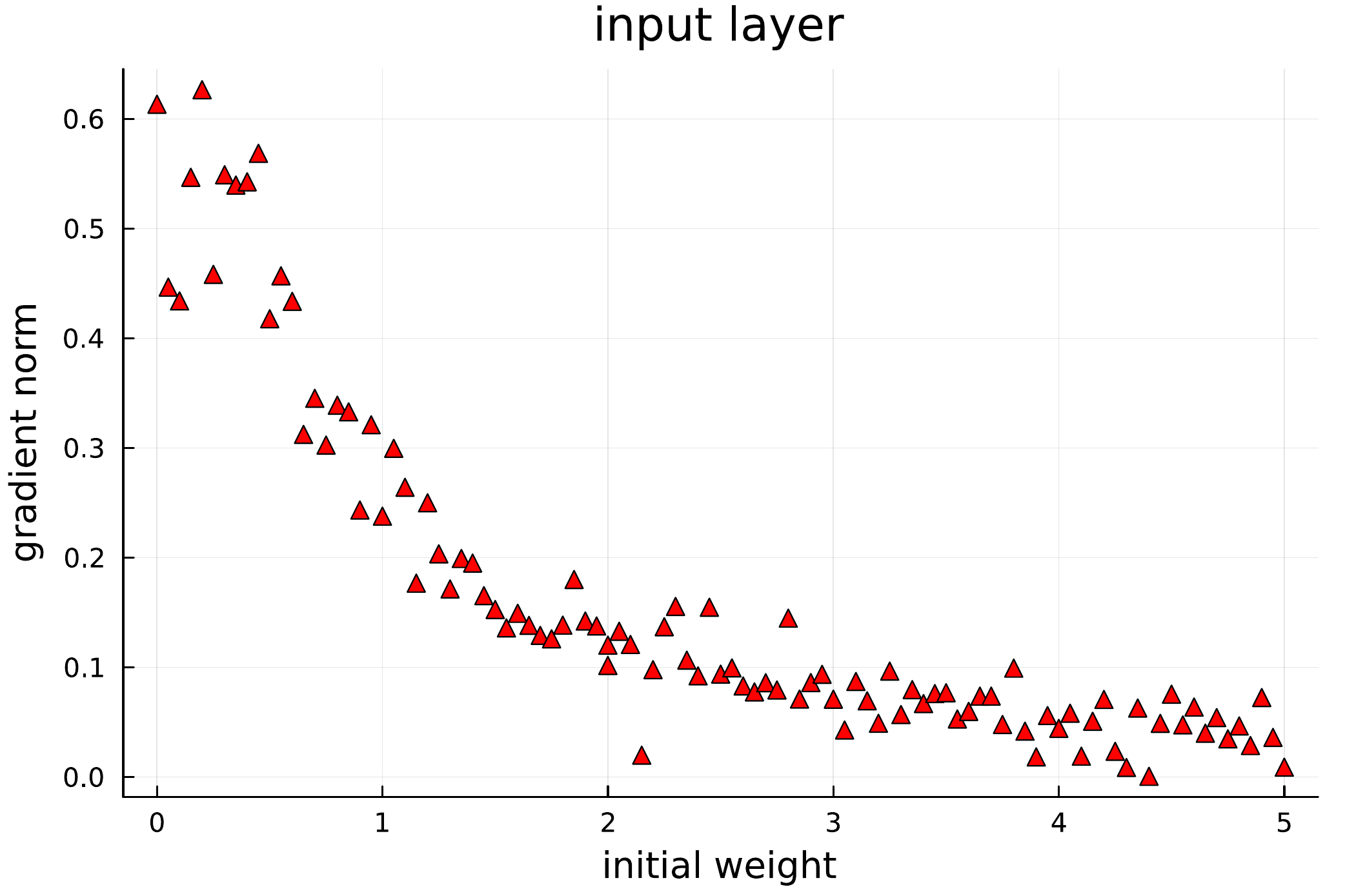}\label{finalgrada}}
\subfloat[][]{\includegraphics[width=0.30\textwidth]{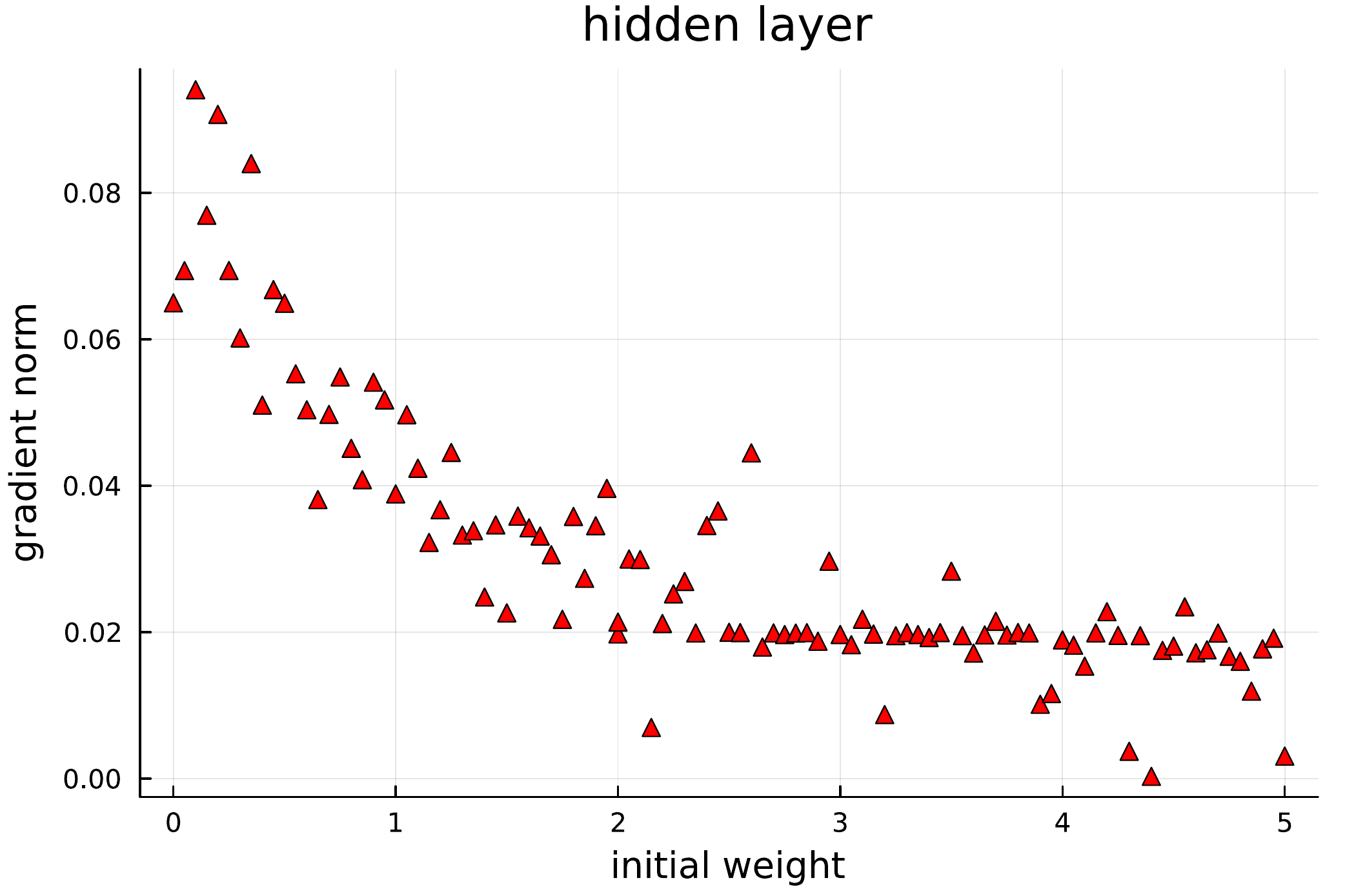}\label{finalgradb}}
\subfloat[][]{\includegraphics[width=0.30\textwidth]{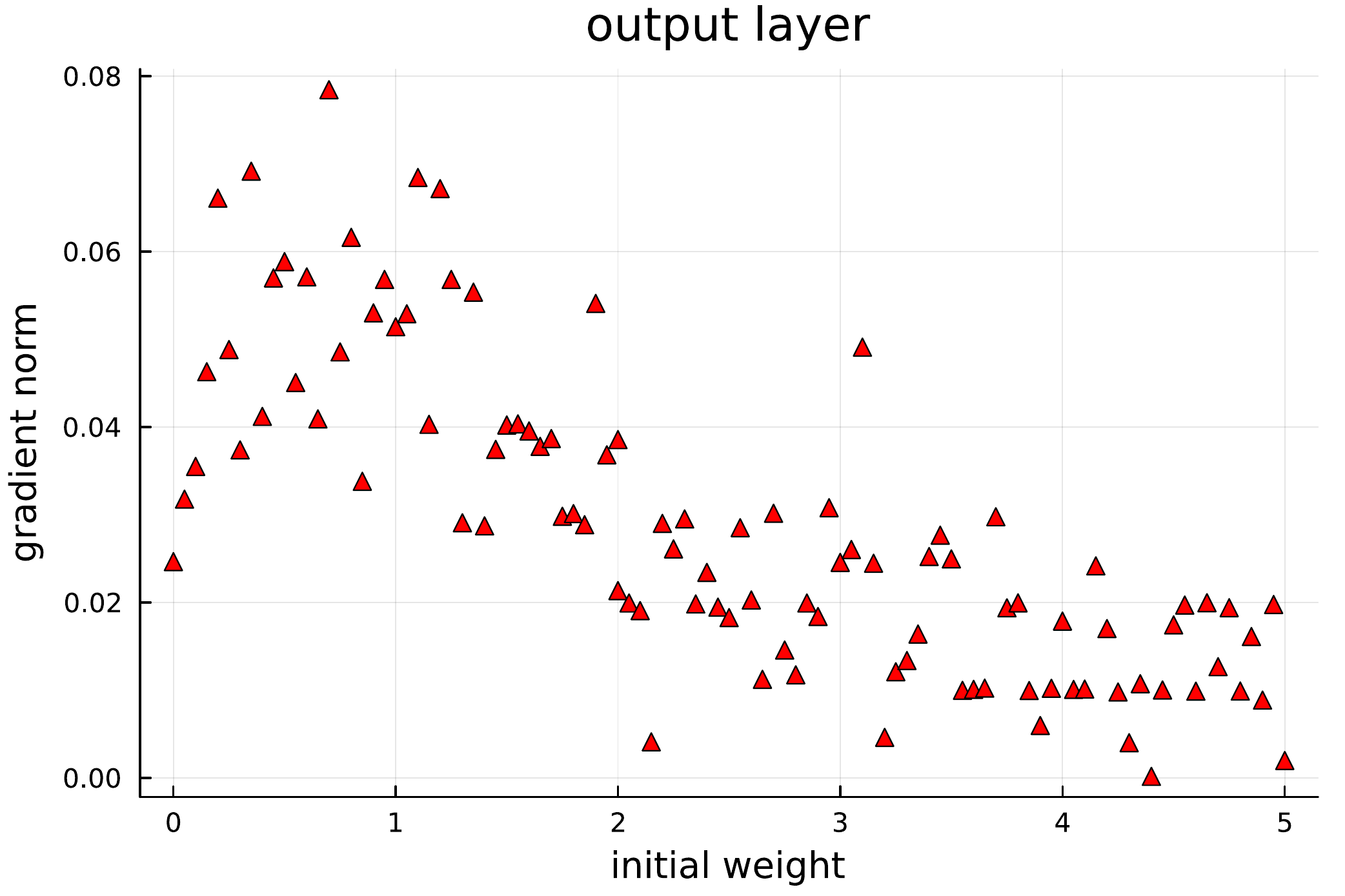}\label{finalgradc}}\\
\caption{
\protect\subref{initialgrada}-\protect\subref{initialgradc}  The infinite norm of the gradient at the first update step as a function of the initial value scale. \protect\subref{finalgrada}-\protect\subref{finalgradc} Same, but  the norm of the gradient after 3 epochs. Note the different scale on the vertical axis, showing how the norm of the gradients has reduced over time. Each point corresponds to a single simulation. } 
\label{initialvalues}
\end{figure}
\par
These results are consistent with the view that rapid feedback alignment during early updates is  important for the eventual performance of the algorithm. A closer examination, however, reveals some additional complexities, which provide further insight. The first one is highlighted by fig.~\ref{initialvalues}, which shows the norm of the gradient for the input, hidden and output layers after the first update step, so after the algorithm has been presented with the first example of the training set (figs.~\ref{initialgrada}-\ref{initialgradc}). The main observation to be made from these figures is that the norm  reduces rapidly  as the initial weights increase.  If we take the norm as an indicator for the step size of the random walker, then  this suggests that walkers initialised with high weights suffer from slow speed as a result of small update steps. High weights, therefore mean an effectively reduced learning rate at the beginning of learning.  Note the dramatic decrease of the learning rate as the  weights start to differ from 0.  
\par
%
%
%
%
\begin{figure}
\subfloat[][]{\includegraphics[width=0.3\textwidth]{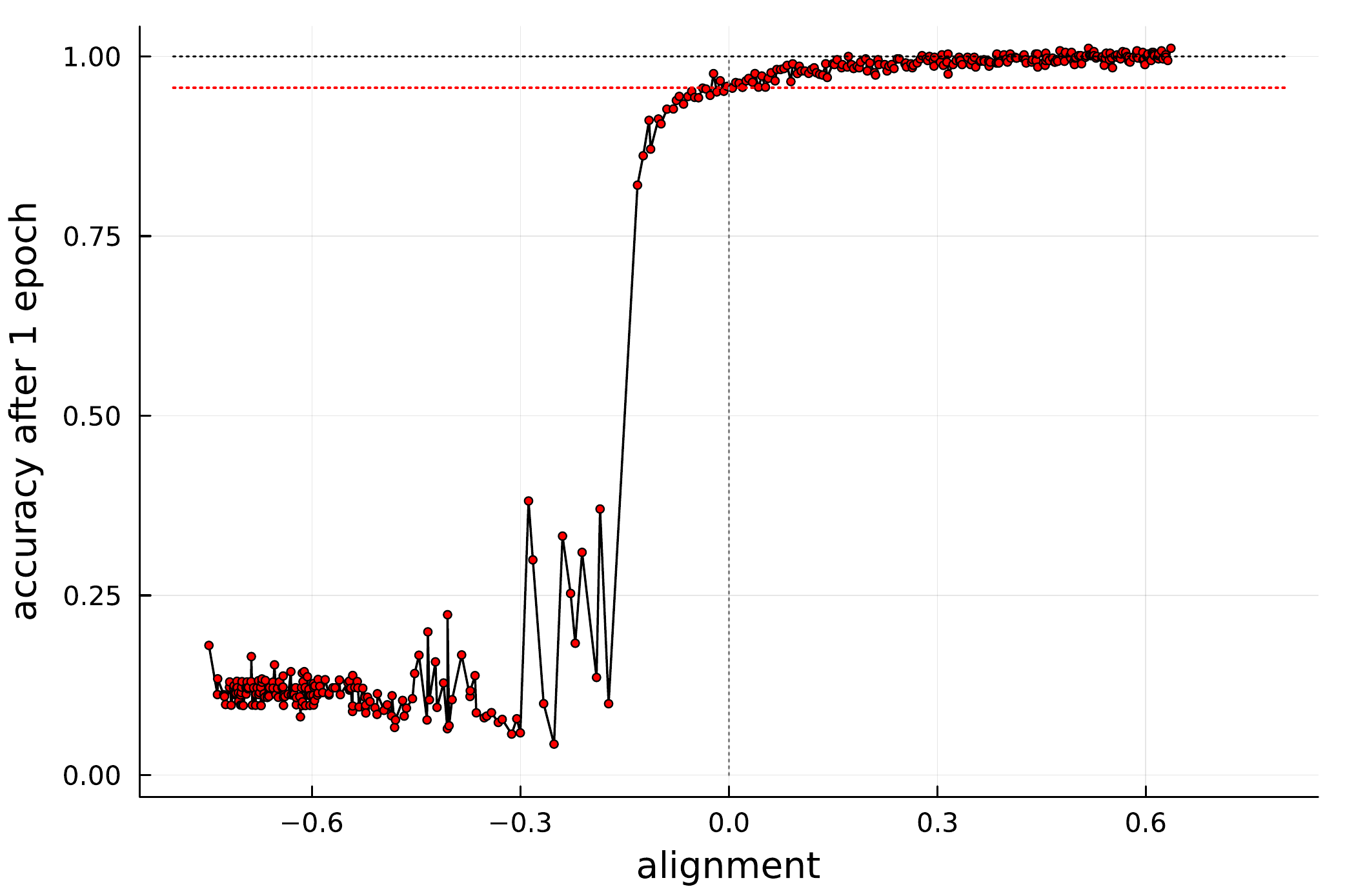}\label{randomdisturb2}}
\subfloat[][]{\includegraphics[width=0.3\textwidth]{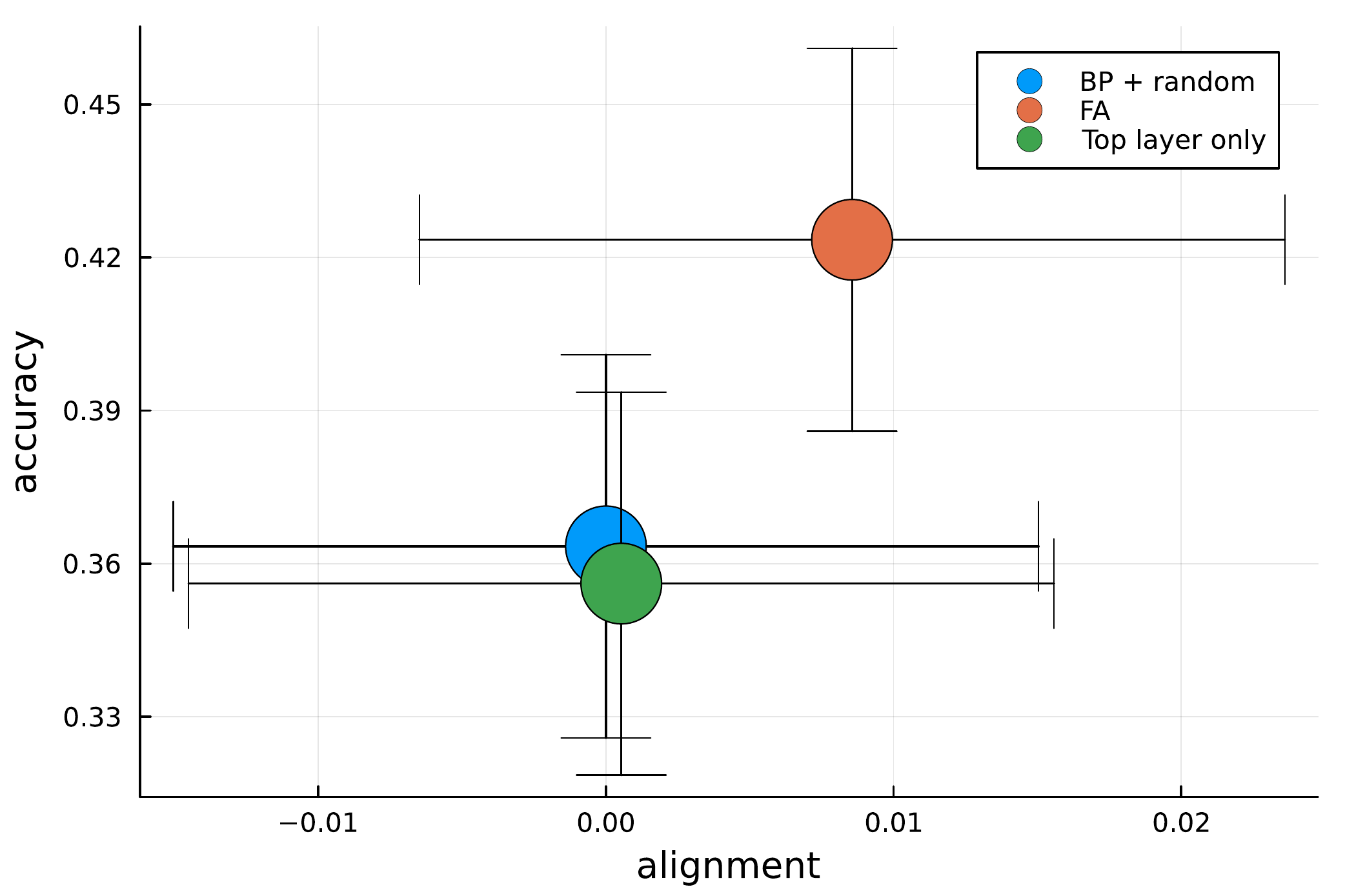}\label{randomdisturbcompare}}
\subfloat[][]{\includegraphics[width=0.3\textwidth]{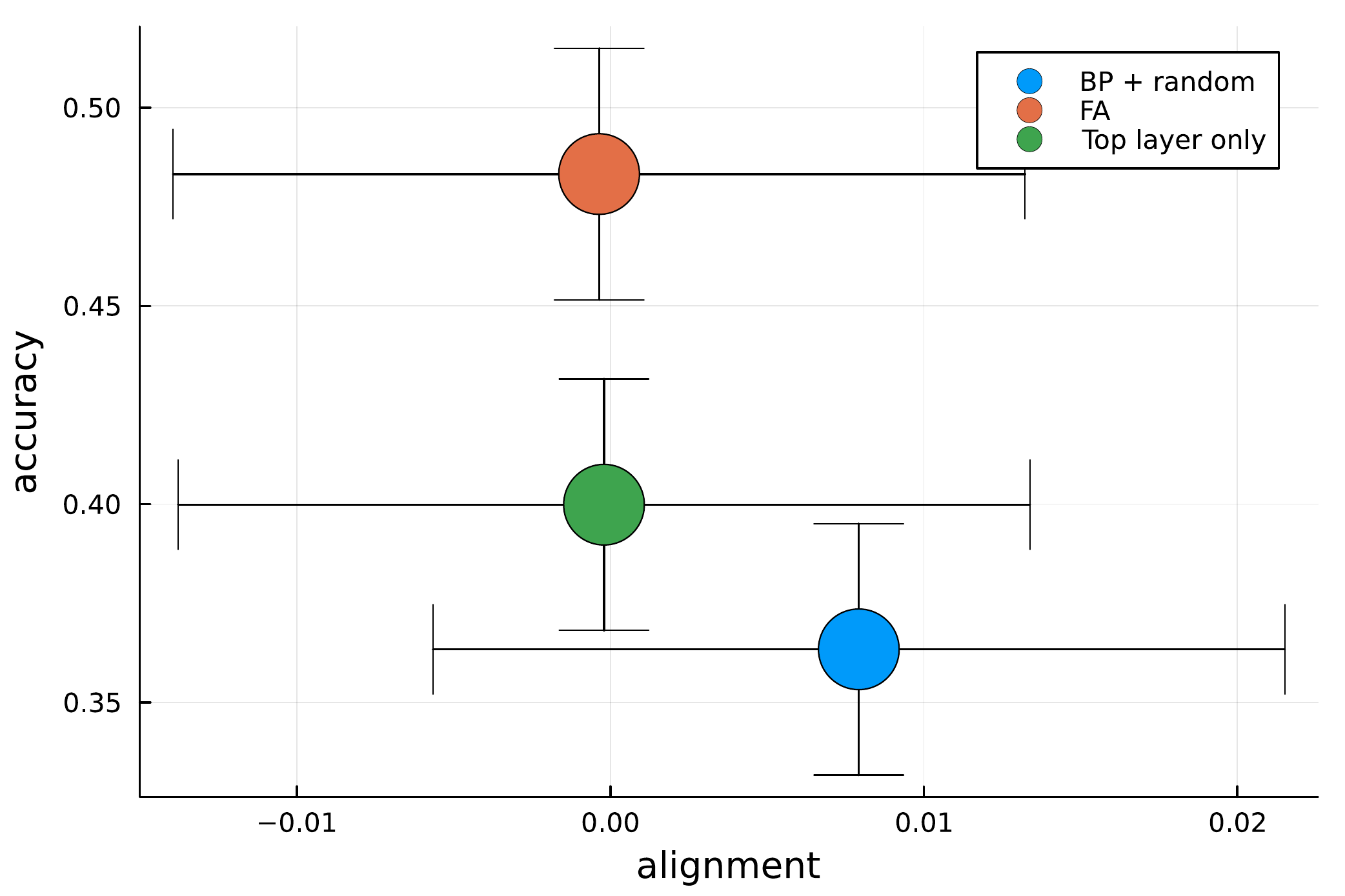}\label{randomdisturbcompare4}}
\caption{  \protect\subref{randomdisturb2} We trained a gradient using BP. For the input and hidden  layer we perturbed the gradient so that it had an angle relative to the actual gradient. The graph shows the accuracy after one epoch relative to BP. A value of 1 indicates that the perturbed network performs equally well as BP. The red line indicates the performance of a network where only the last layer is trained. The accuracy after 5 updates against the mean gradient alignment for the \protect\subref{randomdisturbcompare} input and \protect\subref{randomdisturbcompare4} hidden layer. The average is taken over 500 repetitions. The error bars show the standard deviations. For comparison, the perturbed BP is also shown. Clearly, FA does better than the perturbed BP in those examples.}
\label{randomdisturb}
\end{figure} 
This suggests a new explanation for the improved performance of networks initialised with low weights: FA with initially vanishing weights may perform better because the initial speed of exploration is faster, and hence  accuracy can increase over fewer update steps.  This means that gradient alignment is not necessarily the only reason why FA performs. At least, we have shown one example which suggests a different explanation. We hasten to add that the two explanations are not mutually exclusive. 
\par
This begs now the question whether or not gradient alignment drives accuracy, or whether gradient alignment is merely a by-product of  the FA dynamics. This is best explored during the earliest stages of learning, before a substantial alignment has formed.  In order to  be able to investigate this, we first  need to understand the relationship between gradient alignment and loss. To this end, we generated a baseline curve as follows:  We used  the  BP algorithm to train the  network for  1 epoch on the MNIST dataset. However,   each time the gradient was computed in the hidden and input layer, we randomly perturbed it, such that the actual gradient used for updating the weights was different from the gradient determined by BP. We could then,  for each experiment, determine the alignment between the true gradient and the perturbed gradient. We did this systematically in fig.~\ref{randomdisturb2}, which shows the accuracy after 1 epoch as a function of the average alignment. By design, this set of experiments  isolates the effect of the gradient alignment, while all other details of the algorithm are left the same. The figure also shows in red, the baseline of a multi-layer network where only the last layer is trained, whereas all other layers remain at the initial weights. 
\par
A number of observations can be made. Firstly, the accuracy of the network intersects the red line for an alignment of 0. In this case,  the perturbed gradients  are orthogonal to the actual gradients one would obtain from  BP which means that  weight  updates  do not have an overall drift either into the direction of better accuracy or away from it. Consistently,  the accuracy of the network then corresponds to simulations where only the output  layer is trained (indicated by the red line in fig.~\ref{randomdisturb2}). Further increasing the alignment, only improves the performance a little bit, reflecting the  well known fact that training only the output  layer  leads to high accuracies in many cases. On the other hand, for negative alignments, the performance quickly drops to random guessing, which   corresponds to an  accuracy of $0.1$. (Note that the effect of updating against the gradient is a randomisation of the performance, rather than guiding the network to an accuracy below $0.1$, which would require updating the weights of the output layer against the gradient as well.) A further observation to be made from this is that the gradient descent process operating at the output layer cannot settle on a minimum when the weights in the lower layers have a systematic drift in one direction.    
 \par
Fig.~\ref{randomdisturb2} shows how alignment (or rather misalignment) with the BP gradient impacts  learning performance.  We can now use this  to see whether the  FA performance is driven  solely by the alignment of the pseudo-gradient  or  whether  there is more to it. If  FA performs exactly as well as the perturbed BP with the same alignment, then we know that gradient alignment is driving performance. If, on the other hand, it performs better, then there is some additional  driver.    
\par
Fig.~\ref{randomdisturbcompare} gives some relevant insight. It  shows the performance as a function of the  average alignment for   FA, gradient perturbed BP and  and  a network   where only the top layer is trained. Fig.~\protect\ref{randomdisturbcompare} and \protect\ref{randomdisturbcompare4} compare the performance of the three algorithms after   5 updates against the alignment of the  input  and hidden layer. This shows that, for similar alignment, FA does much better than the other two algorithms.  This suggests two things: \one Gradient  alignment   is not the only driver of  performance of FA, at least not during early stages. \two The performance of the FA during early stages is not entirely driven by the gradient descent in the output layer, but the update of the input and hidden layer does add to performance.  

\subsubsection{Alignment can reduce performance}

So far, we have established that alignment between the gradient and pseudo-gradient is not driving performance during early stages of learning. We will now show that alignment is not sufficient for performance, even during later stages, and indeed can be outright detrimental. 
\par
\begin{figure}
\subfloat[][]{\includegraphics[width=0.30\textwidth]{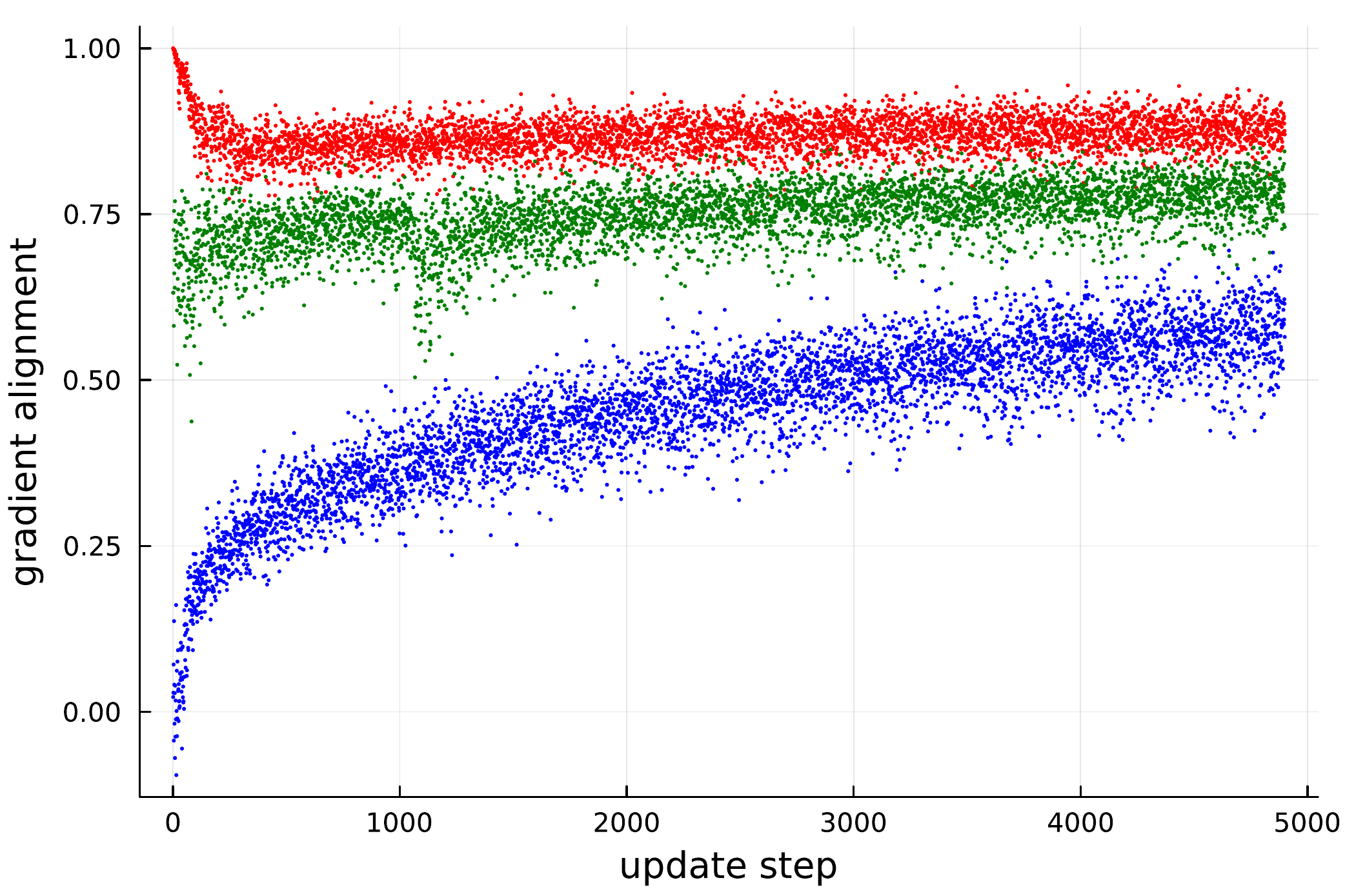}\label{angle}}
\subfloat[][]{\includegraphics[width=0.30\textwidth]{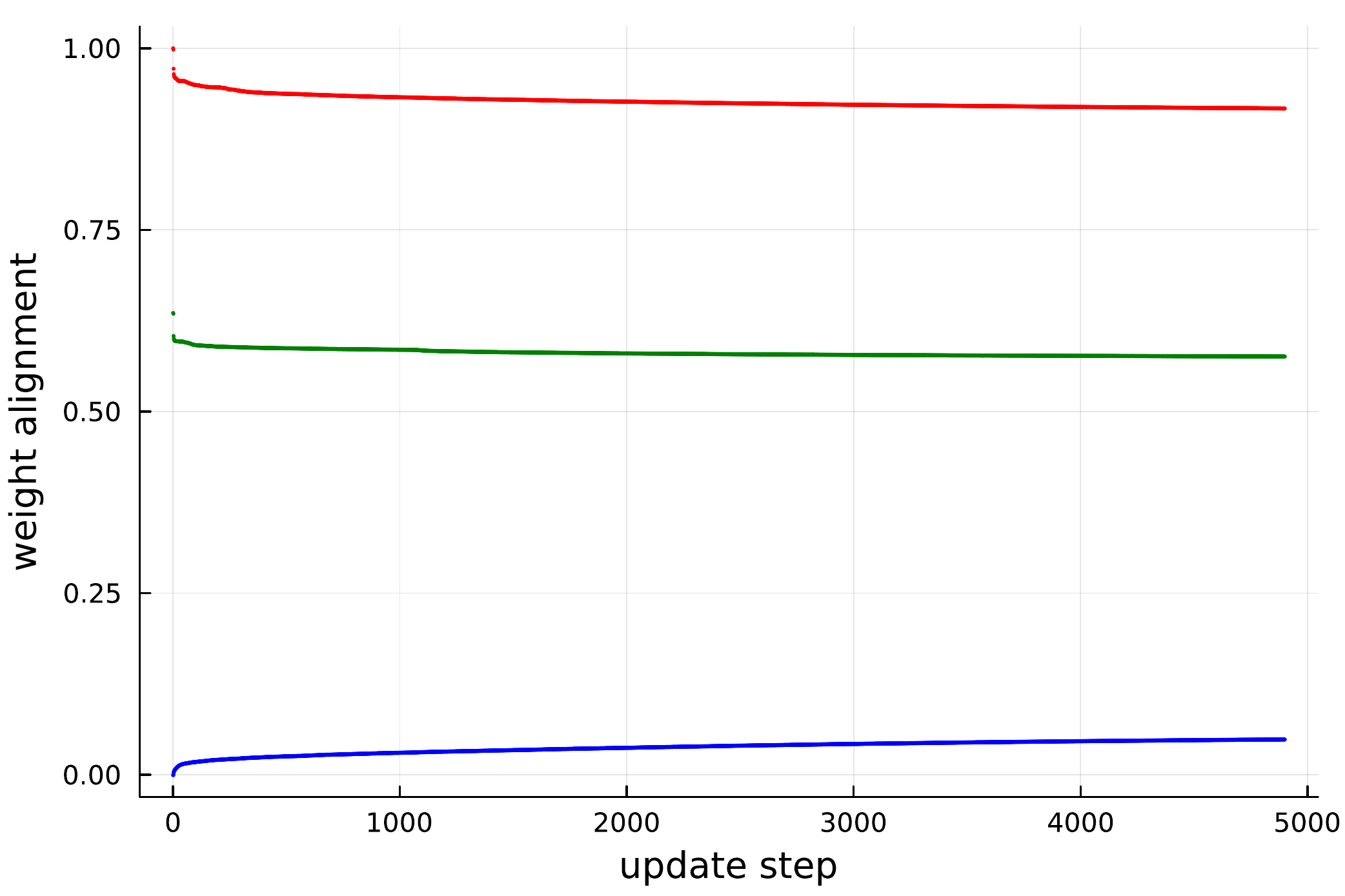}\label{weight}}
\subfloat[][]{\includegraphics[width=0.30\textwidth]{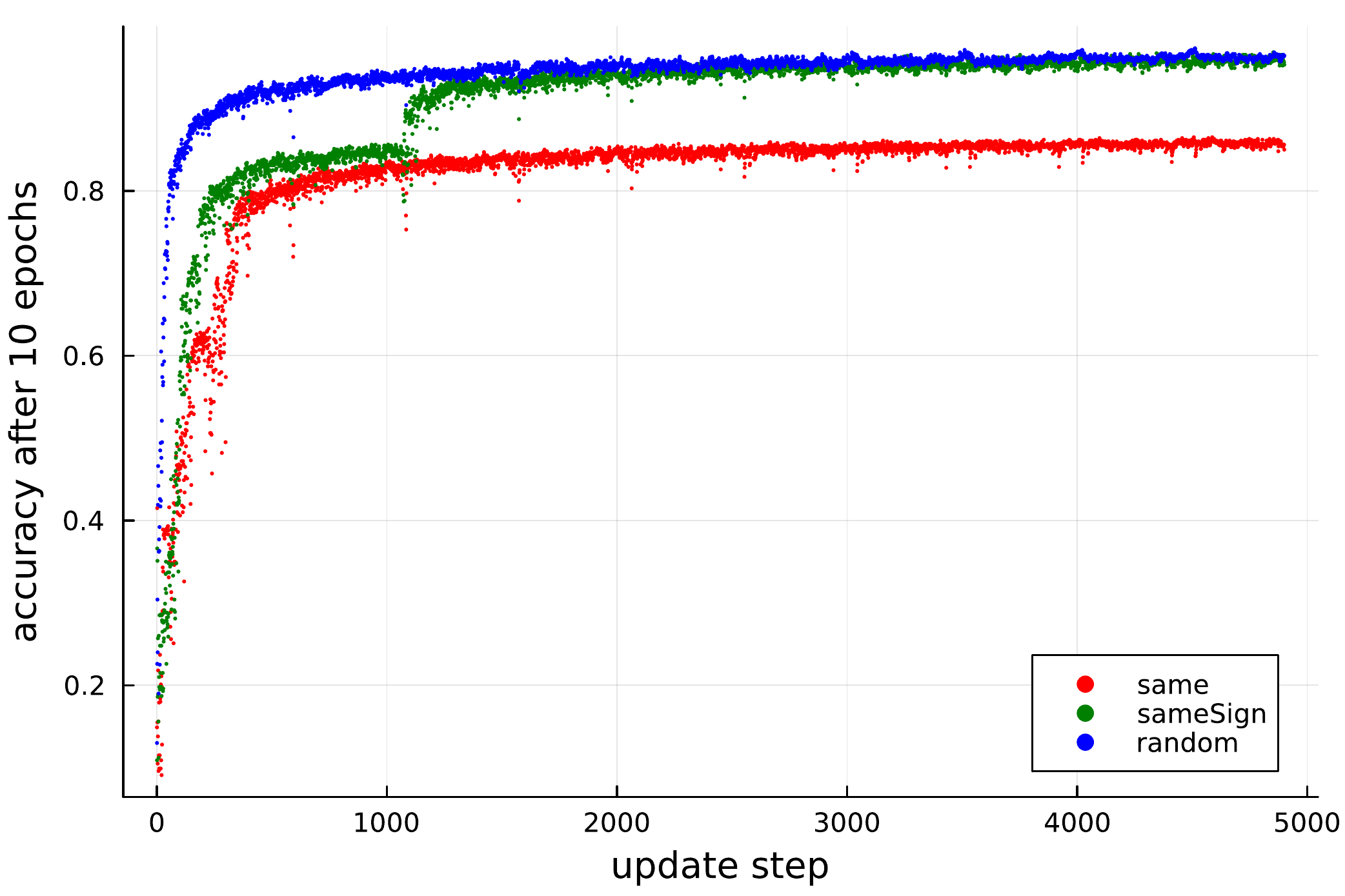}\label{accuracy}}
\caption{Feedback alignment, with initial weights chosen at random (blue), randomly but the sign of each element was the same as the feedback matrix (green) and exactly the same as the feedback matrix (red). We show  \protect\subref{angle}  the angle between the true gradient and the feedback gradient of the hidden layer, \protect\subref{angle} the alignment between the weights of the hidden layer and the feedback matrix, \protect\subref{accuracy}  the accuracy over time. There is no trend between higher alignment and better accuracy, or better increase of accuracy. Each point  represents a value taken after an update step. Altogether, the simulations here represent 10 epochs. } 
\label{inverseexample}
\end{figure}

Fig.~\ref{inverseexample} shows, as an example, a set of three  different simulations  of  FA with  identical hyper-parameters. The  only difference between them is  the initialisation of the weights.   The blue line is just a standard simulation, with weights being initially drawn from a normal distribution before being scaled by $0.05$.   The green simulation is the same, but the signs of the initial weights  were set equal to the entries of the corresponding  feedback matrices. As a consequence, there is a large initial alignment between the  weights and the feedback matrices. Finally, the red points  show the results for a simulation where  the initial weights were set to be identical to the feedback matrices before being  scaled  by $0.05$. For all three simulations, we  drew  the elements of the feedback matrices from a normal distribution, rather than from the set $\{-1,1\}$. We did this so that the initial weights in the blue simulation are statistically indistinguishable from the other two simulations, while the initial alignment between the weights is different in the three cases. By construction, the blue simulation is unaligned initially, the red simulation is perfectly aligned and the green simulation is somewhere in between those two cases.

%
\par
We find from the simulations  presented in  fig.~\ref{inverseexample}, as expected, that good weight alignment translates to high  gradient alignment. The key message of the graph is that the  initially unaligned simulation performs best, amongst the three runs. While we are not claiming that high gradient alignment is always detrimental, from this single example we can still draw  two conclusions: \one Weight alignment is not a necessary condition for performance of the FA algorithm. \two High gradient alignment is not sufficient for performance of the FA algorithm. 
\par
While the performance of the red simulation is still good in the sense that it is apparently learning something, it is possible to construct examples   of FA networks that have almost perfect alignment throughout, but learn nothing (data not shown). The simplest way to do this is to  use feedback matrices that are initialised by randomly drawing elements from the set  $\{-1, 1\}$, and set the initial weights equal to the feedback matrices. These initial conditions are not conducive to algorithm performance, and the network does not train well. Altogether, we find,  that  gradient alignment is not always sufficient to explain the performance of FA algorithms. We could show at least one example, where other  mechanisms are required. 

\section{Discussion}

At a first glance, FA and DFA should not work. By replacing a key term in the update equation,  the gradient descent of BP is effectively transformed into a random walk in weight-space. The key observation that  had been made  early on was that this random walk aligns with the ``true'' BP gradient. In the literature this alignment is commonly assumed to be the driver for the performance of the FA algorithm. It remains unclear, however, why FA aligns. Existing mathematical models only cover special simplified cases of neural networks. They suggest that the update dynamics of FA leads to weight alignment, which then implies gradient alignment. In that sense, FA approximates the true gradient.  However, weight alignment is typically much weaker than gradient alignment. This suggests that some other explanations are required. 
\par
Our theoretical results suggest a somewhat different view:  FA is not  approximating BP at all, and indeed does not descend or ascend the  gradient of a loss function or an approximation  thereof.    It is not ``learning'' in the sense one normally understands this term.  Instead, it  performs a random walk in weight space.  It works based on the following conjectured mechanism:
At the output layer, the gradient descend process drives the network as a whole towards a fixed point, corresponding to a vanishing gradient of the loss function. Many of those fixed points are unstable for the random walker and the updates in the hidden and input layer will drive the network away from the fixed point, until a fixed point is found for which the stability criterion is valid. Then, the network will converge towards this fixed point. 
\par
There is no guarantee that FA finds extrema that are compatible with the alignment criterion. Failure to do so  could be either because there are no compatible extrema, or because  FA is initialised in a part of parameter space from which it cannot find a route to compatible fixed points. The latter scenario could realise when extrema of the loss function are sparse in parameter space or when  the weights are initialised in an area that has small update steps, for example  for large initial weights. The often cited failure of FA for convolutional neural networks is likely a consequence of the sparseness of loss extrema for those networks. 
\par
Throughout this article we concentrated on FA, but  clearly the conclusions can be transferred directly to DFA. The only difference between FA and DFA is that  in the latter  each layer performs an independent walk, whereas in  FA the training process is a single random walker in a  higher dimensional space. The  basic mechanism of how DFA finds extrema  remains the same as in the case of FA. 
\par
Some open questions remain. The  first one relates to the distribution of local extrema of the loss function in parameter space. In particular, it may be that for certain types of problems there are conditions that guarantee that FA and DFA find local extrema or alternatively, that they do not find such extrema (as seems to be the case for convolutional neural networks). There is also a lack of theoretical understanding of spatially inhomogeneous random walks. In order to come to a complete theoretical description  of FA, we need a sound justification of how and under which conditions random walks approach fixed points.

\section{Methods}

Unless otherwise stated, we used  a feed forward multi-layer perceptron with an input/hidden layer of 700/1000 neurons. The size of the network was chosen to be large, while still being fast to execute. Our results are not sensitive to variations of the size of the network, although classification performance clearly is.   Throughout, we used $\tanh$ as the activation function; we also experimented with {\tt relu} which led to qualitatively the same results (data not shown). For the final layer we used the softmax function and as a loss function we used the cross-entropy.   The batch size was chosen to be 100 and the learning rate was set to $0.05$. Again, our results are not sensitive to those parameters.
\par
Unless stated otherwise, feedback matrices were chosen randomly by drawing matrix elements from the set $\{-1,1\}$. We found FA not to be  overly sensitive to the particular choice of this set. However, we found algorithm performance to  depend on it to some extent. The particular choice we made gives good performance, but we made no attempt to choose the optimal  one. 
\par
For all layers, initial weights were drawn from a normal distribution and then scaled with a weight scaling factor, resulting in both negative and positive weights. If the factor is zero, then all weights were initially zero. The larger the factor, the larger the (absolute value of) initial weights. 
\par
All simulations were done using Julia (1.8.5) Flux for gradient computations and network construction, and CUDA for simulations on GPUs. Throughout no optimisers were used. Weight updates were made by adding the gradient, scaled by the learning rate, to the weights. The networks were trained on the MNIST dataset as included with the Flux library. Whenever an accuracy is reported it was computed based on the test-set of the MNIST dataset.


 \appendix

\section{Two basic properties of FA}
\label{basicproperties}

\subsection{Vanishing weights are unstable for some activation functions}
As pointed out by \cite{refinetti}, for  BP the initialisation $w_{ij}=0$  results in a deadlock, in the sense that  irrespective of the inputs, the weight updates will always vanish. This can be seen clearly from eq.~\ref{nextequation}: For each layer, the matrices $\partial h_i^{(l)}/\partial f_j^{(l)}$   evaluate to $w_{ij}^{(l)}$. Therefore, trivially the weight updates will be vanishing as well.  In the case of FA, these  expressions are replaced by non-vanishing random terms. With this change, the  update of the $l=1$ layer, %
\begin{displaymath}
\Delta w_{pq} = \partial\mathcal L_i B_{ij}^{(1)} f_{jp}^{(1)} x_q
\end{displaymath}
is no longer proportional to the weights, and hence not identically zero. Upon the  first update, the weights of the input layer will typically be non-vanishing, such that the update of the $l=2$ layer will be non-vanishing as well. After $L$ updates, all layers may have weights that are different from 0. The initialisation $w_{ij}=0$ is  therefore not a fixed points of the FA pseudo-update rule.

\subsection{Weight increments tend to zero for large weights for some choices of activation function}

The behaviour for large weights depends on the activation functions that are chosen.  It is therefore not possible to make general statements. However,  many popular choices for these functions will lead to vanishing weight updates. We restrict the discussion to the choices made here.  The derivative of the  softmax function  $\mathbf\sigma(x_1,\ldots,x_n)$  can be seen to vanish at infinite weights. The derivative of the first component of the softmax function with respect to $x_1$ gives
\begin{displaymath}
{\partial \sigma_1 \over \partial x_1}= {\exp(x_1)\over \sum_i \exp(x_i} - \left({\exp(x_1)\over \sum_i \exp(x_i)}\right)^2.
\end{displaymath}
It is straightforward to confirm that as   $\max(x_i)\rightarrow\infty$ the first and the second term simultaneously either go to 0 or 1. Similarly, for $k\neq 1$ 
\begin{displaymath}
{\partial \sigma_1 \over \partial x_k}= - {\exp(x_1 )\over \sum_i \exp(x_k)}{\exp(x_i )\over \sum_i \exp(x_i)}.
\end{displaymath}
Here at least one of the two terms vanishes for as long as the difference between any two $x_j$ is sufficiently large.  
\par
From the shape of the   $\tanh(x)$ it is clear that its derivative vanishes   with $x\rightarrow \pm\infty$. The same is not true, however, for relu which has a constant derivative.
\par
We conclude that for networks that use the softmax function weight updates will vanish in the regime of large weights when the softmax function is used in the last layer.

\end{document}